\documentclass[letterpaper,10pt,conference]{ieeeconf}
\IEEEoverridecommandlockouts
\let\labelindent\relax
\usepackage[utf8]{inputenc} %
\usepackage[T1]{fontenc}    %
\usepackage{graphicx} \graphicspath{ {figures/} }
\usepackage{amsmath,amssymb,mathabx,amsbsy,mathtools,etoolbox}
\usepackage{times}
\usepackage{graphicx}
\usepackage{amsmath,amssymb,mathbbol}
\usepackage{acronym}
\usepackage{enumitem}
\usepackage[breaklinks,colorlinks,linkcolor=black]{hyperref}
\usepackage{balance}
\usepackage{xspace}
\usepackage{setspace}
\usepackage[skip=3pt,font=small]{subcaption}
\usepackage[skip=3pt,font=small]{caption}
\usepackage[dvipsnames,svgnames,x11names]{xcolor}
\usepackage[capitalise,noabbrev,nameinlink]{cleveref}
\usepackage{booktabs,tabularx,colortbl,multirow,multicol,array,makecell}
\usepackage{dblfloatfix}
\usepackage[misc]{ifsym}
\usepackage{pifont}
\usepackage{cite}

\makeatletter
\DeclareRobustCommand\onedot{\futurelet\@let@token\@onedot}
\def\@onedot{\ifx\@let@token.\else.\null\fi\xspace}

\def\etc{\emph{etc}\onedot}

\def\wrt{w.r.t\onedot} 

\def\etal{\emph{et al}\onedot}
\makeatother

\makeatletter
\def\BState{\State\hskip-\ALG@thistlm}
\makeatother

\makeatletter
\renewcommand{\paragraph}{%
  \@startsection{paragraph}{4}%
  {\z@}{0ex \@plus 0ex \@minus 0ex}{-1em}%
  {\hskip\parindent\normalfont\normalsize\bfseries}%
}
\makeatother

\crefname{algorithm}{Alg.}{Algs.}
\Crefname{algocf}{Algorithm}{Algorithms}
\crefname{section}{Sec.}{Secs.}
\Crefname{section}{Section}{Sections}
\crefname{table}{Tab.}{Tabs.}
\Crefname{table}{Table}{Tables}
\crefname{figure}{Fig.}{Fig.}
\Crefname{figure}{Figure}{Figure}

\definecolor{gblue}{HTML}{4285F4}
\definecolor{gred}{HTML}{DB4437}
\definecolor{ggreen}{HTML}{0F9D58}

\definecolor{mygray}{gray}{.92}

\usepackage{pifont}
\usepackage{bbm}
\usepackage{soul}
\usepackage[export]{adjustbox}

\newcommand{\method}{GenDexGrasp\xspace}
\newcommand{\nhand}{5\xspace}
\newcommand{\nobj}{58\xspace}
\newcommand{\ngrasp}{436,000\xspace}
\newcommand{\dataset}{MultiDex\xspace}

\newcommand{\qH}{q_{H}}
\newcommand{\HH}{\mathcal{H}}
\newcommand{\OO}{\mathcal{O}}
\newcommand{\dfc}{\textsc{dfc}\xspace}
\newcommand{\mala}{\textsc{mala}}
\newcommand{\CC}{\mathcal{C}}
\newcommand{\DD}{\mathcal{D}}
\newcommand{\hhO}{\hat{\hat{\Omega}}}

\title{\LARGE \bf \method: Generalizable Dexterous Grasping}

\author{Puhao Li$^{1,2,\star}$, Tengyu Liu$^{1,\star}$, Yuyang Li$^{1,2}$, Yiran Geng$^{1,3}$, Yixin Zhu$^{3}$, Yaodong Yang$^{1,3}$, Siyuan Huang$^{1,\dagger}$%
\thanks{$^{\star}$ Puhao Li and Tengyu Liu contributed equally to this paper.}%
\thanks{$^\dagger$ Corresponding email: {\tt{syhuang@bigai.ai}}.}
\thanks{$^{1}$ Beijing Institute of General Artificial Intelligence (BIGAI).}%
\thanks{$^{2}$ Tsinghua University. \quad{} $^{3}$ Peking University.}%
}

\let\oldtwocolumn\twocolumn
\renewcommand\twocolumn[1][]{%
    \oldtwocolumn[{#1}{
        \vspace{-21pt}
        \centering
        \includegraphics[width=\textwidth]{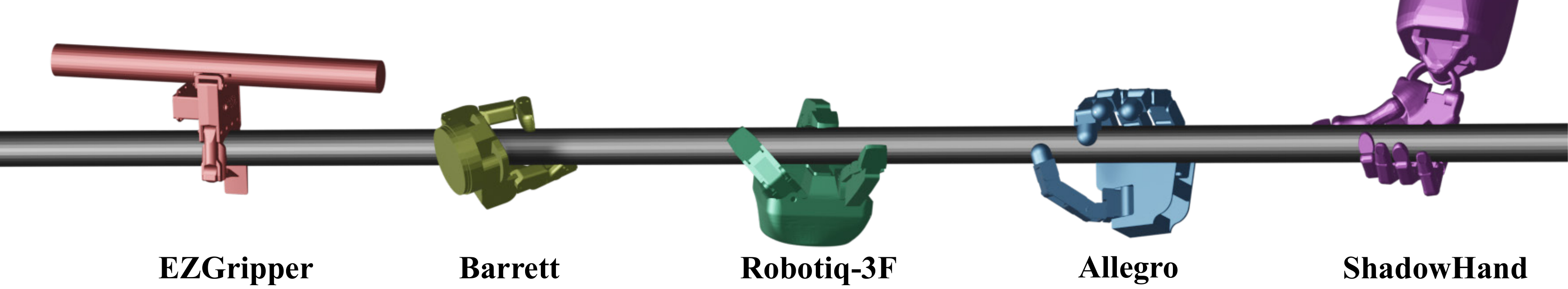}
        \captionof{figure}{\textbf{The proposed \method synthesizes and generalizes versatile dexterous grasps across arbitrary robot hands.}}
        \label{fig:teaser}
    }]
}

\begin{document}
\maketitle
\thispagestyle{empty}
\pagestyle{empty}

\begin{abstract}
Generating dexterous grasping has been a long-standing and challenging robotic task. Despite recent progress, existing methods primarily suffer from two issues. First, most prior art focuses on a specific type of robot hand, lacking \textit{generalizable} capability of handling unseen ones. Second, prior arts oftentimes fail to \textit{rapidly} generate \textit{diverse} grasps with a \textit{high success rate}. To jointly tackle these challenges with a unified solution, we propose the \method, a novel hand-agnostic grasping algorithm for generalizable grasping. \method is trained on our proposed large-scale multi-hand grasping dataset \dataset synthesized with force closure optimization. By leveraging the contact map as a hand-agnostic intermediate representation, \method efficiently generates diverse and plausible grasping poses with a high success rate and can transfer among diverse multi-fingered robotic hands. Compared with previous methods, \method achieves a three-way trade-off among success rate, inference speed, and diversity.\let\thefootnote\relax\footnotetext{Code: \url{https://github.com/tengyu-liu/GenDexGrasp}.}
\end{abstract}

\setstretch{0.95}

\section{Introduction}

Humans' ability to grasp is astonishingly versatile. In addition to the full grasp with five fingers, humans can efficiently \textit{generalize} grasps when some fingers are occupied and imagine \textit{diverse} grasping poses for various downstream tasks when given an unseen new type of hand, all happened \textit{rapidly} with \textit{a high success rate}. These criteria starkly contrast with most prior robot grasping methods, which primarily focus on specific end-effectors, requiring redundant efforts to learn the grasp model for every new robotic hand. On top of this challenge, prior methods often have difficulties quickly generating diverse hand poses for unseen scenarios, further widening the gap between robot and human capabilities. Hence, these deficiencies necessitate a generalizable grasping algorithm, efficiently handling arbitrary situations and allowing fast prototyping for new robots.

Fundamentally, the most significant challenge in generalizable dexterous grasping~\cite{arunachalam2022dexterous,dexmanip,mandikal2022dexvip,jiang2021hand,radosavovic2021state,kokic2020learning,wu2022learning} is to find an \textit{efficient} and \textit{transferable} representation for diverse grasp. The \emph{de facto} representation, joint angles, is unsuitable for its dependency on the structure definition: two similar robotic hands could have contrasting joint angles if their joints are defined differently. Existing works use contact points~\cite{liu2021synthesizing,shao2020unigrasp,li2022efficientgrasp}, contact maps~\cite{brahmbhatt2019contactgrasp,turpin2022graspd}, and approach vectors~\cite{xu2021adagrasp} as the representations, and execute the desired grasps with complex solvers. A simple yet effective representation is still in need.

In this paper, we denote \textbf{generalizable dexterous grasping} as the problem of generating grasping poses for unseen hands. We evaluate generalizable grasping in three aspects:
\begin{itemize}[leftmargin=*,noitemsep,nolistsep]
    \item \textbf{Speed:} Hand-agnostic methods adopt inefficient sampling strategies~\cite{brahmbhatt2019contactgrasp,liu2021synthesizing,turpin2022graspd}, which leads to extremely slow grasp generation, ranging from 5 minutes to 40 minutes.
    \item \textbf{Diversity:} Hand-aware methods~\cite{shao2020unigrasp,xu2021adagrasp,li2022efficientgrasp} rely on deterministic solvers, either as a policy for direct execution or predicted contact points for inverse kinematics, resulting in identical grasping poses for the same object-hand pair.
    \item \textbf{Generalizability:} Hand-aware methods~\cite{shao2020unigrasp,xu2021adagrasp,li2022efficientgrasp} also rely on hand descriptors trained on two- and three-finger robotic hands, which hinders their generalizability to new hands that are drastically different from the trained ones.
\end{itemize}

To achieve a three-way trade-off among the above aspects and alleviate the aforementioned issues, we devise \method for generalizable dexterous grasping. Inspired by Brahmbhatt \etal~\cite{brahmbhatt2019contactgrasp}, we first generate a hand-agnostic contact map for the given object using a conditional variational autoencoder~\cite{sohn2015learning}. Next, we optimize the hand pose to match the generated contact map. Finally, the grasping pose is further refined in a physics simulation to ensure a physically plausible contact. \method provides generalizability by reducing assumptions about hand structures and achieves fast inference with an improved contact map and an efficient optimization scheme, resulting in diverse grasp generation by a variational generative model with random initialization.

To address contact ambiguities (especially for thin-shell objects) during grasp optimization, we devise an aligned distance to compute the distance between surface point and hand, which helps to represent accurate contact maps for grasp generation. Specifically, the traditional Euclidean distance would mistakenly label both sides of a thin shell as contact points when the contact is on one side, whereas the aligned distance considers directional alignment to the surface normal of the contact point and rectifies the errors.

To learn the hand-agnostic contact maps, we collect a large-scale multi-hand dataset, \dataset, using force closure optimization~\cite{liu2021synthesizing}. \dataset contains \ngrasp diverse grasping poses for \nhand hands and \nobj household objects.

\setstretch{1}

We summarize our contributions as follows:
\begin{enumerate}[leftmargin=*,noitemsep,nolistsep]
    \item We propose \method, a versatile generalizable grasping algorithm. \method achieves a three-way trade-off among speed, diversity, and generalizability to unseen hands. We demonstrate that \method is significantly faster than existing hand-agnostic methods and generates more diversified grasping poses than hand-aware methods. Our method also achieves strong generalizability, comparable to existing hand-agnostic methods. 
    \item We devise an aligned distance for properly measuring the distance between the object's surface point and hand. We represent a contact map with the aligned distance, which significantly increases the grasp success rate, especially for thin-shell objects. The ablation analysis in \cref{tbl:ablation-contact} shows the efficacy of such a design.
    \item We collect and open-source a large-scale synthetic dataset, \dataset, for generalizable grasping with \nhand robotic hands, \nobj household objects, and \ngrasp diverse grasping poses. \dataset is by far the largest multi-hand grasp dataset with diverse hand structures.
\end{enumerate}

\section{Related Work}

\subsection{Generalizable Dexterous Grasping}

Existing solutions to generalizable grasping fall into two categories: hand-aware and hand-agnostic. 
The hand-aware methods are limited by the diversity of generated poses, whereas the hand-agnostic methods are oftentimes too slow for various tasks. Below, we review both methods in detail.

\textit{Hand-aware} approaches~\cite{shao2020unigrasp,xu2021adagrasp,li2022efficientgrasp} learn a data-driven representation of the hand structure and use a neural network to predict an intermediate goal, which is further used to generate the final grasp. For instance, UniGrasp~\cite{shao2020unigrasp} and EfficientGrasp~\cite{li2022efficientgrasp} extract the gripper's PointNet~\cite{qi2017pointnet} features in various poses and use a PSSN network to predict the contact points of the desired grasp. As a result, contact points are used as the inverse kinematics's goal, which generates the grasping pose. Similarly, AdaGrasp~\cite{xu2021adagrasp} adopts 3D convolutional neural networks to extract gripper features, ranks all possible poses from which the gripper should approach the object, and executes the best grasp with a planner. However, all hand-aware methods train and evaluate the gripper encoders only with two- and three-finger grippers, hindering their ability to generalize to unseen grippers or handle unseen scenarios. Critically, these methods solve the final grasp deterministically, yielding similar grasping poses.

\textit{Hand-agnostic} methods rely on carefully designed sampling strategies~\cite{brahmbhatt2019contactgrasp,liu2021synthesizing,turpin2022graspd}. For instance, ContactGrasp~\cite{brahmbhatt2019contactgrasp} leverages the classic grasp planner in \textit{GraspIt!}~\cite{miller2004graspit} to match a selected contact map, and Liu \etal~\cite{liu2021synthesizing} and Turpin \etal~\cite{turpin2022graspd} sample hand-centric contact points/forces and update the hand pose to minimize the difference between desired contacts and actual ones. All these methods adopt stochastic sampling strategies that are extremely slow to overcome the local minima in the landscape of objective functions. As a result, existing hand-agnostic methods take minutes to generate a new grasp, impractical for real-world applications.

\subsection{Contact Map}

Contact map has been an essential component in modern grasp generation and reconstruction. Initialized by GraspIt!~\cite{miller2004graspit} and optimized by DART~\cite{schmidt2014dart}, ContactGrasp~\cite{brahmbhatt2019contactgrasp} uses thumb-aligned contact maps from ContactDB~\cite{brahmbhatt2019contactdb} to retarget grasps to different hands. ContactOpt~\cite{grady2021contactopt,brahmbhatt2020contactpose} uses an estimated contact map to improve hand-object interaction reconstruction. NeuralGrasp~\cite{khargonkar2022neuralgrasps} retrieves grasping poses by finding the nearest neighbors in the latent space projections of contact maps. Wu \etal~\cite{wu2022learning} samples contact points on object surfaces and uses inverse kinematics to solve the grasping pose. Mandikal \etal~\cite{mandikal2021learning} treats contact maps as object affordance and learns an RL policy that manipulates the object based on the contact maps. \dfc \cite{liu2021synthesizing} simultaneously updates hand-centric contact points and hand poses to sample diverse and physically stable grasping from a manually designed Gibbs distribution. GraspCVAE~\cite{jiang2021hand} and Grasp'D~\cite{turpin2022graspd} use contact maps to improve grasp synthesis: GraspCVAE generates a grasping pose and refines the pose \wrt an estimated contact map, whereas Grasp'D generates and refines the expected contact forces while updating the grasping pose. IBS-Grasp~\cite{she2022learning} learns a grasping policy that takes an interaction bisector surface, a generalized contact map, as the observed state.
Compared to prior methods, the proposed \method differs by treating the contact map as the \textit{transferable} and \textit{intermediate} representation for \textit{hand-agnostic} grasping. We use a less restrictive contact map and a more efficient optimization method for faster and more diversified grasp generation; see detailed in \cref{sec:contact_map}.

\begin{figure}
    \centering
    \begin{subfigure}{0.2\linewidth}\hfill
    \includegraphics[width=0.9\linewidth,valign=c]{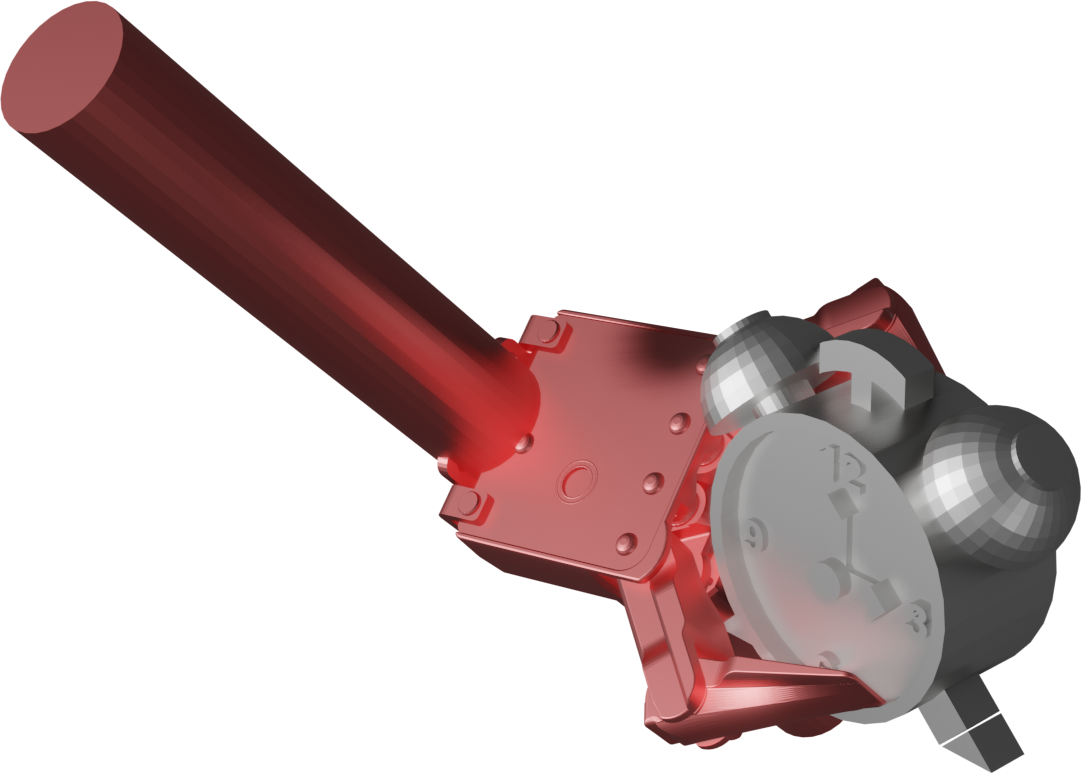} 
    \hfill\end{subfigure} \hfill
    \begin{subfigure}{0.2\linewidth}\hfill
    \includegraphics[width=0.7\linewidth,valign=c]{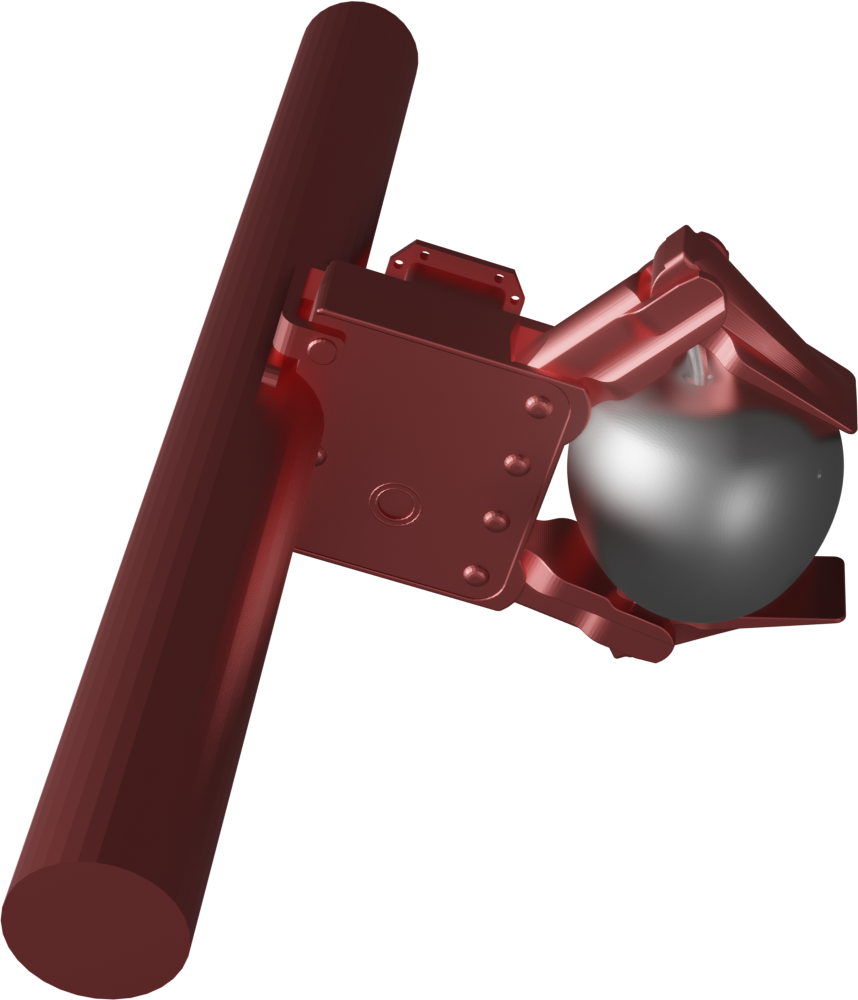} 
    \hfill\end{subfigure} \hfill
    \begin{subfigure}{0.2\linewidth}\hfill
    \includegraphics[width=0.7\linewidth,valign=c]{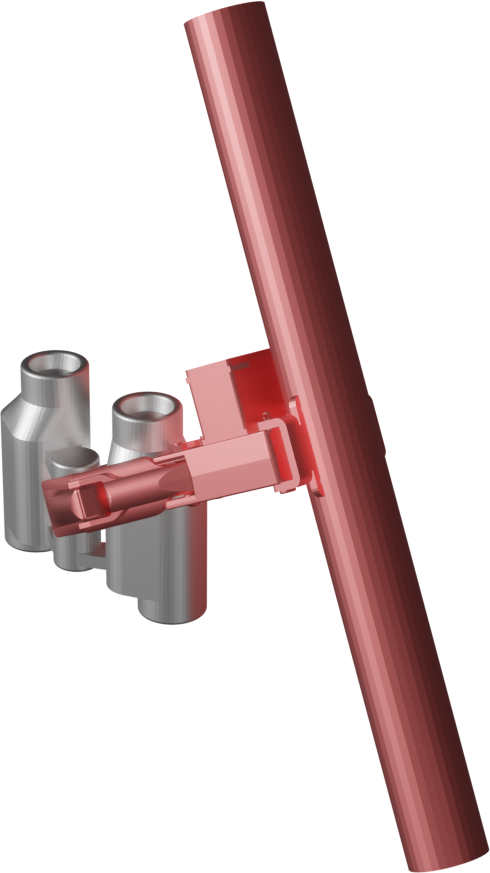} 
    \hfill\end{subfigure} \hfill
    \begin{subfigure}{0.2\linewidth}\hfill
    \includegraphics[width=0.9\linewidth,valign=c]{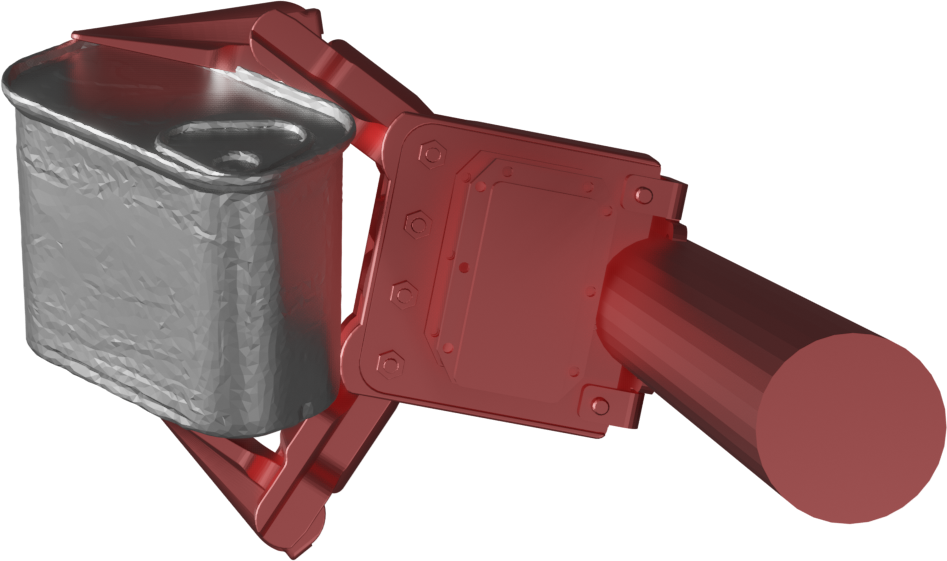} 
    \hfill\end{subfigure} \\\vspace{0.1cm}
    \begin{subfigure}{0.2\linewidth}\hfill
    \includegraphics[width=0.9\linewidth,valign=c]{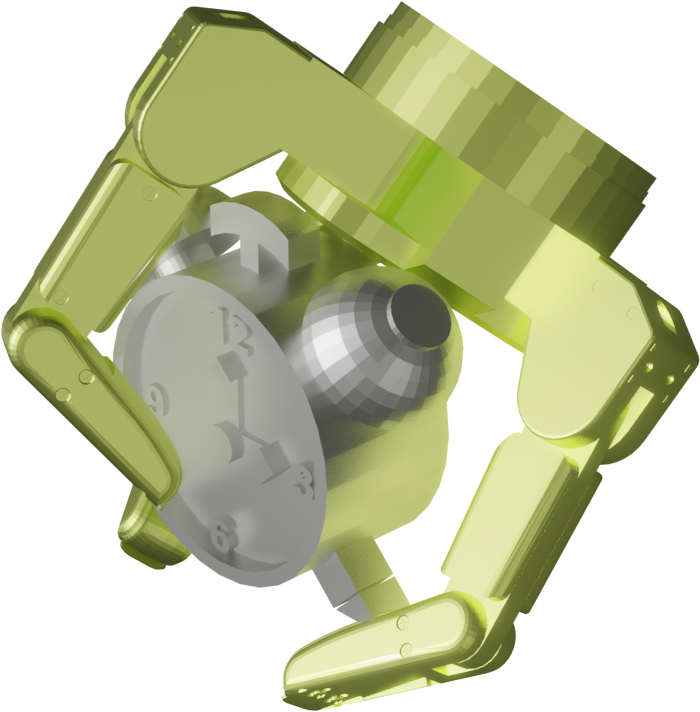} 
    \hfill\end{subfigure} \hfill
    \begin{subfigure}{0.2\linewidth}\hfill
    \includegraphics[width=0.9\linewidth,valign=c]{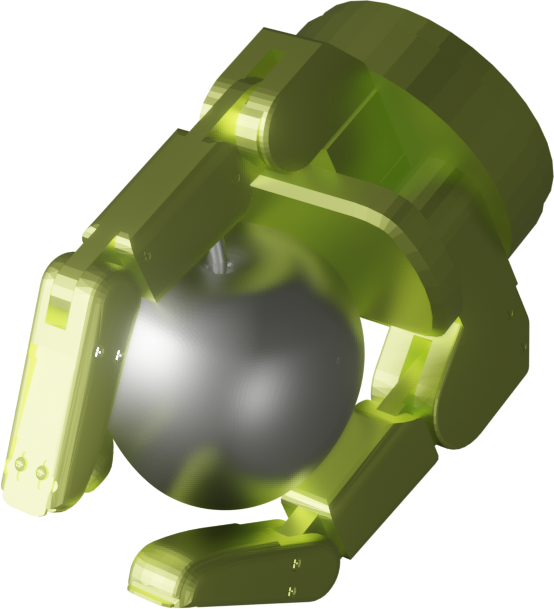} 
    \hfill\end{subfigure} \hfill
    \begin{subfigure}{0.2\linewidth}\hfill
    \includegraphics[width=0.9\linewidth,valign=c]{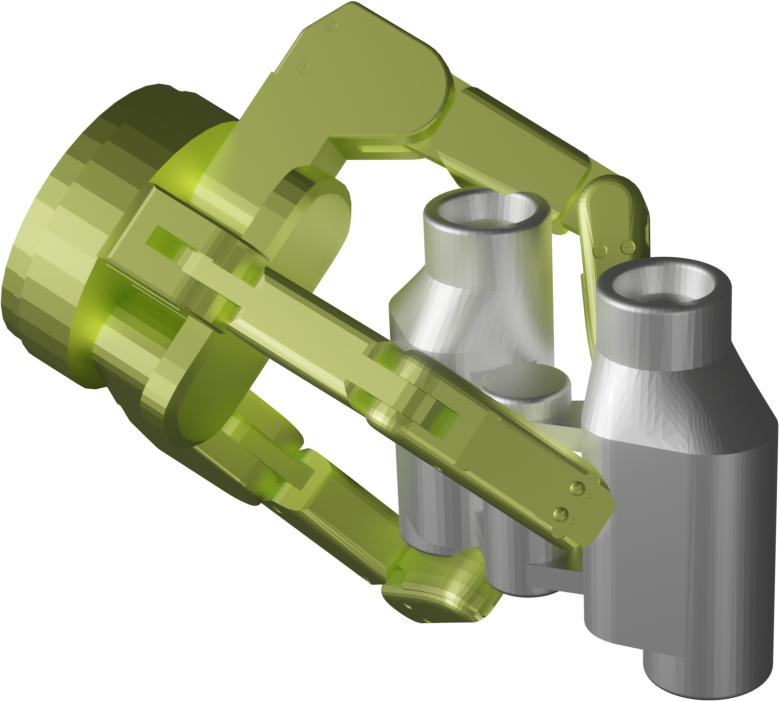} 
    \hfill\end{subfigure} \hfill
    \begin{subfigure}{0.2\linewidth}\hfill
    \includegraphics[width=0.9\linewidth,valign=c]{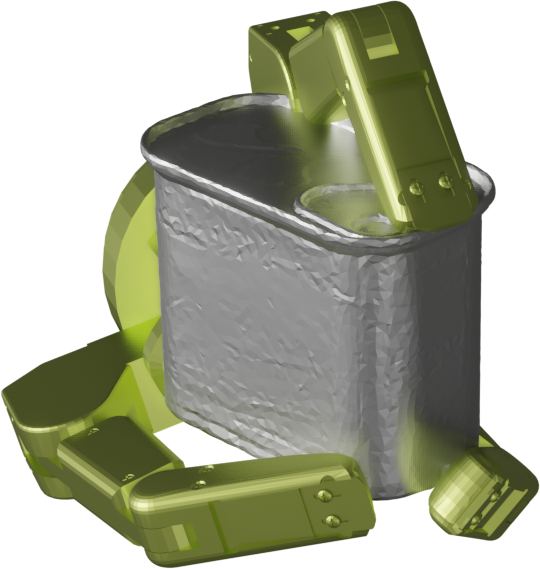} 
    \hfill\end{subfigure} \\\vspace{0.1cm}
    \begin{subfigure}{0.2\linewidth}\hfill
    \includegraphics[width=0.6\linewidth,valign=c]{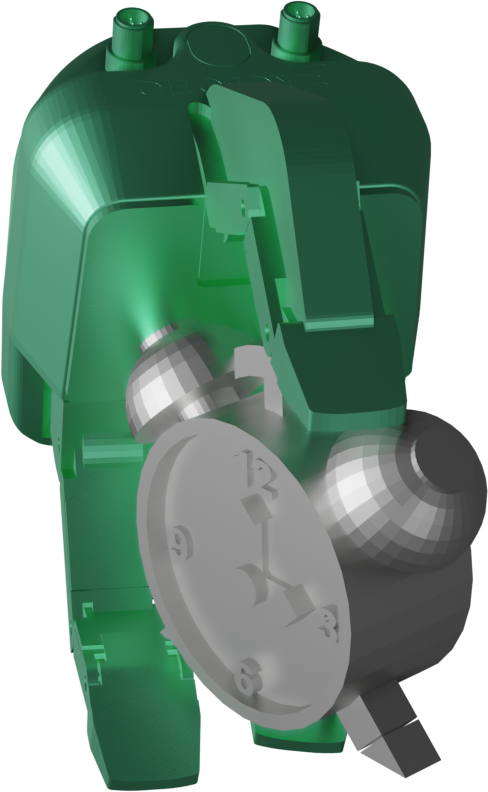}
    \hfill\end{subfigure} \hfill
    \begin{subfigure}{0.2\linewidth}\hfill
    \includegraphics[width=0.9\linewidth,valign=c]{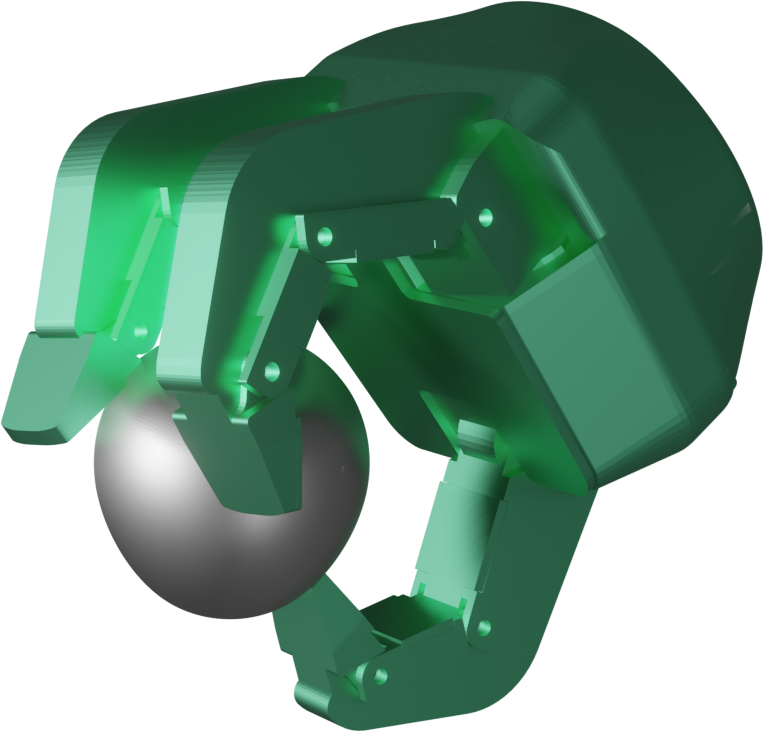} 
    \hfill\end{subfigure} \hfill
    \begin{subfigure}{0.2\linewidth}\hfill
    \includegraphics[width=0.9\linewidth,valign=c]{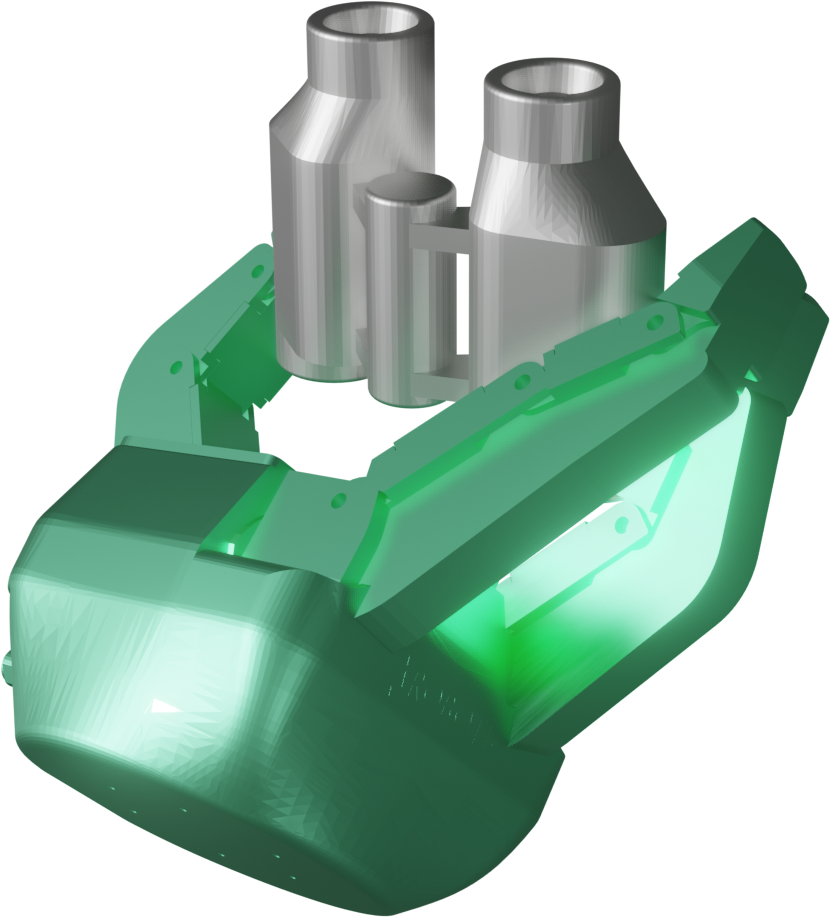} 
    \hfill\end{subfigure} \hfill
    \begin{subfigure}{0.2\linewidth}\hfill
    \includegraphics[width=0.9\linewidth,valign=c]{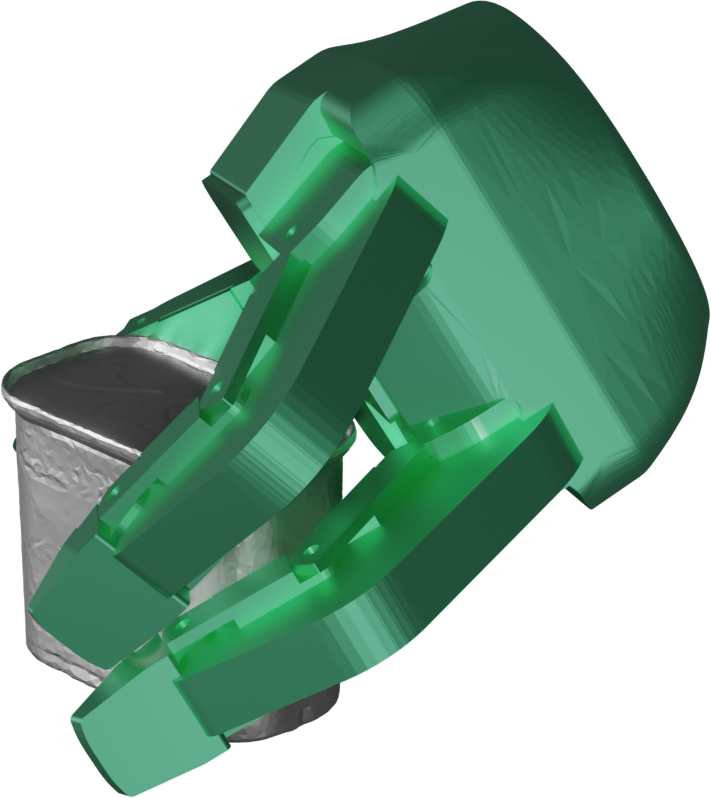} 
    \hfill\end{subfigure} \\\vspace{0.1cm}
    \begin{subfigure}{0.2\linewidth}\hfill
    \includegraphics[width=0.9\linewidth,valign=c]{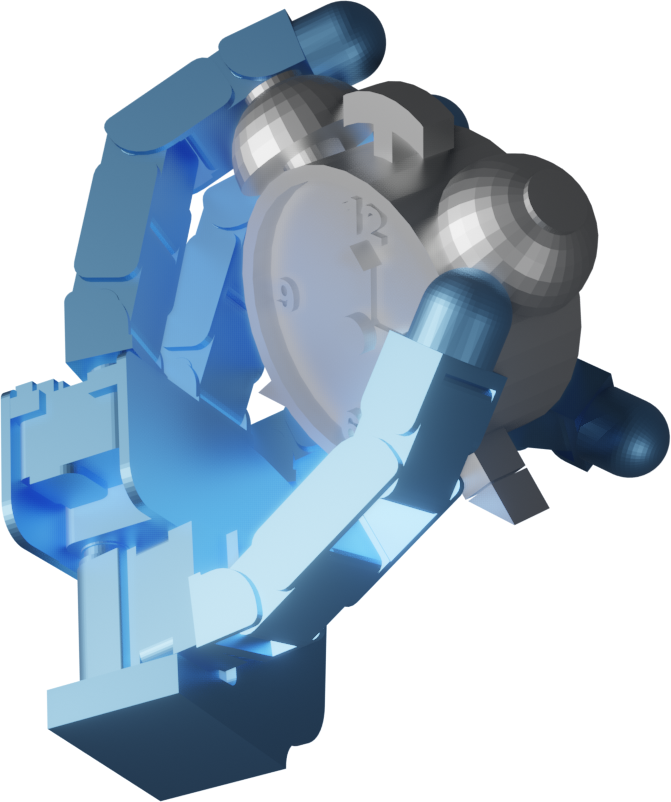} 
    \hfill\end{subfigure} \hfill
    \begin{subfigure}{0.2\linewidth}\hfill
    \includegraphics[width=0.9\linewidth,valign=c]{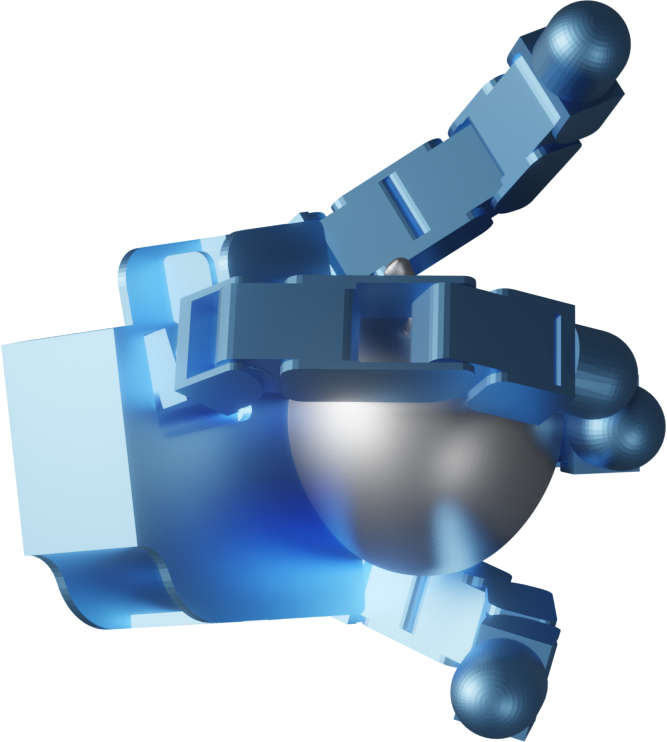} 
    \hfill\end{subfigure} \hfill
    \begin{subfigure}{0.2\linewidth}\hfill
    \includegraphics[width=0.9\linewidth,valign=c]{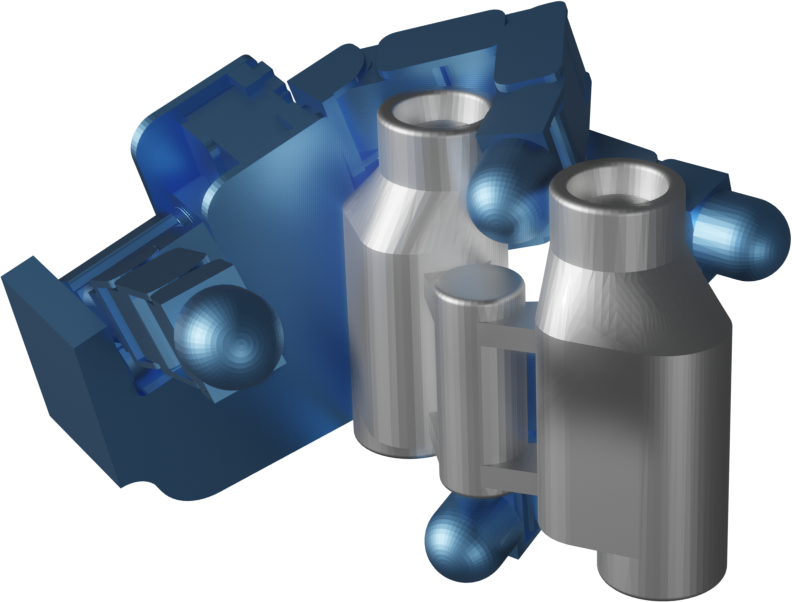} 
    \hfill\end{subfigure} \hfill
    \begin{subfigure}{0.2\linewidth}\hfill
    \includegraphics[width=0.9\linewidth,valign=c]{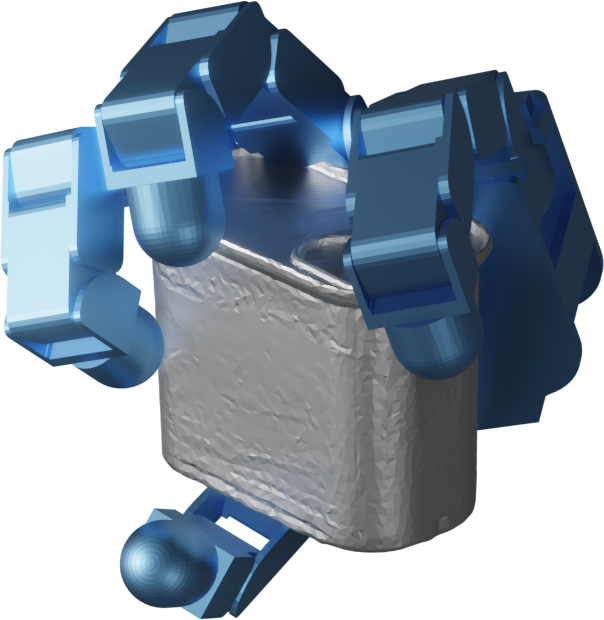} 
    \hfill\end{subfigure} \\\vspace{0.1cm}
    \begin{subfigure}{0.2\linewidth}\hfill
    \includegraphics[width=0.9\linewidth,valign=c]{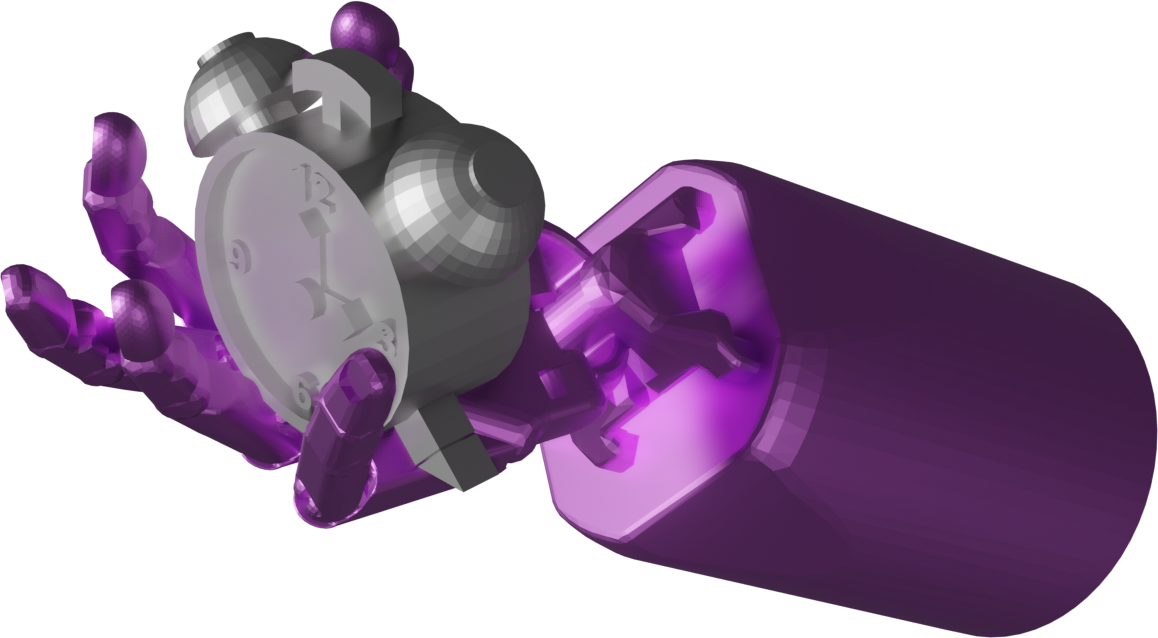}
    \hfill\end{subfigure} \hfill
    \begin{subfigure}{0.2\linewidth}\hfill
    \includegraphics[width=0.9\linewidth,valign=c]{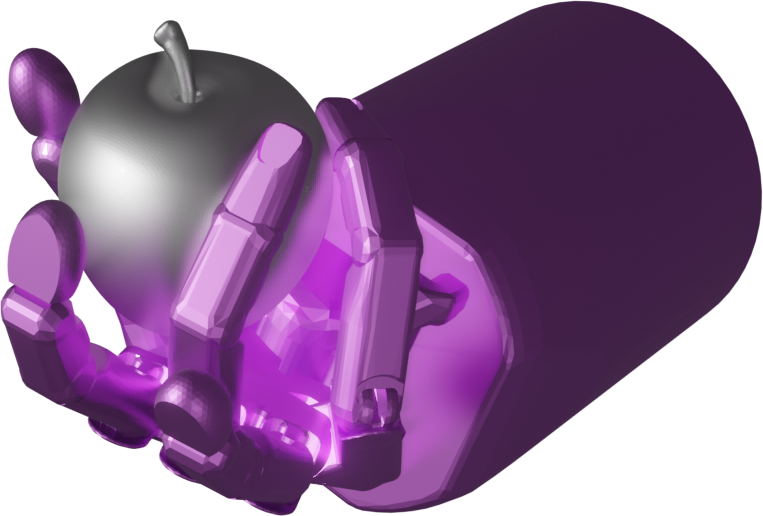}
    \hfill\end{subfigure} \hfill
    \begin{subfigure}{0.2\linewidth}\hfill
    \includegraphics[width=0.9\linewidth,valign=c]{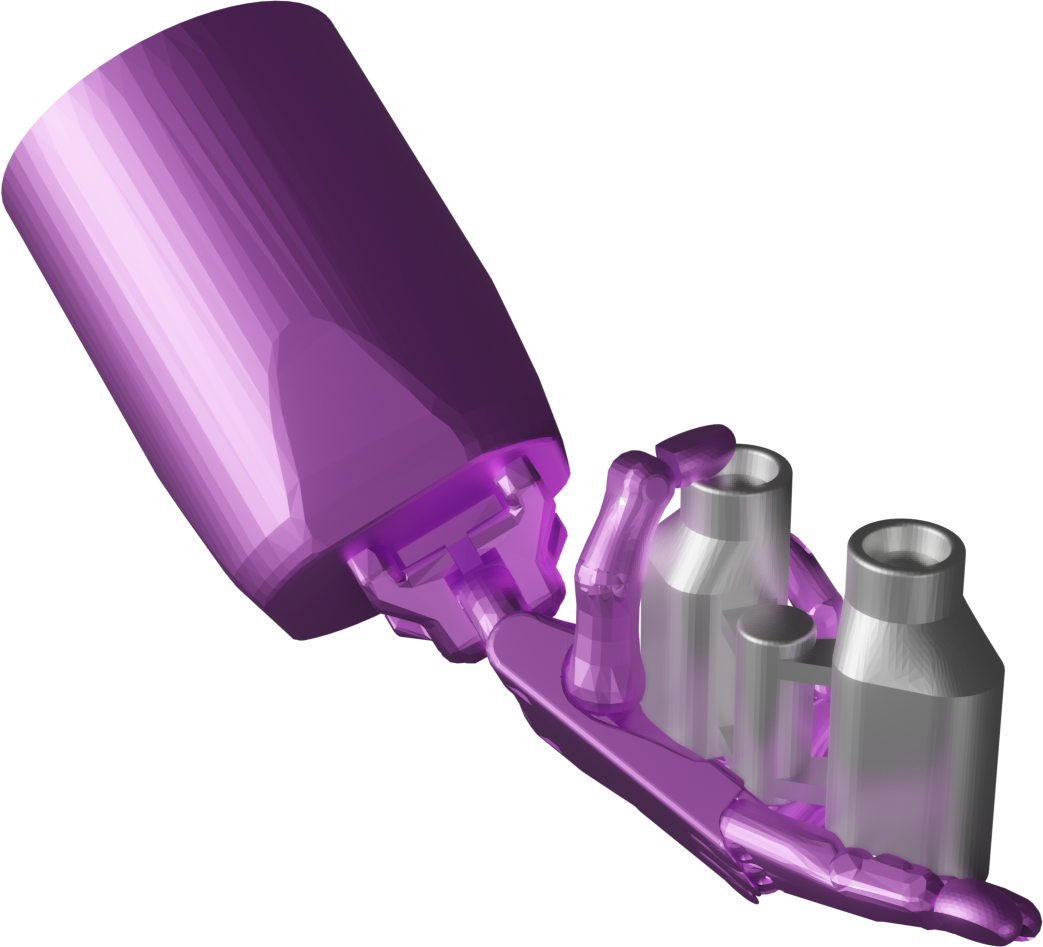}
    \hfill\end{subfigure} \hfill
    \begin{subfigure}{0.2\linewidth}\hfill
    \includegraphics[width=0.9\linewidth,valign=c]{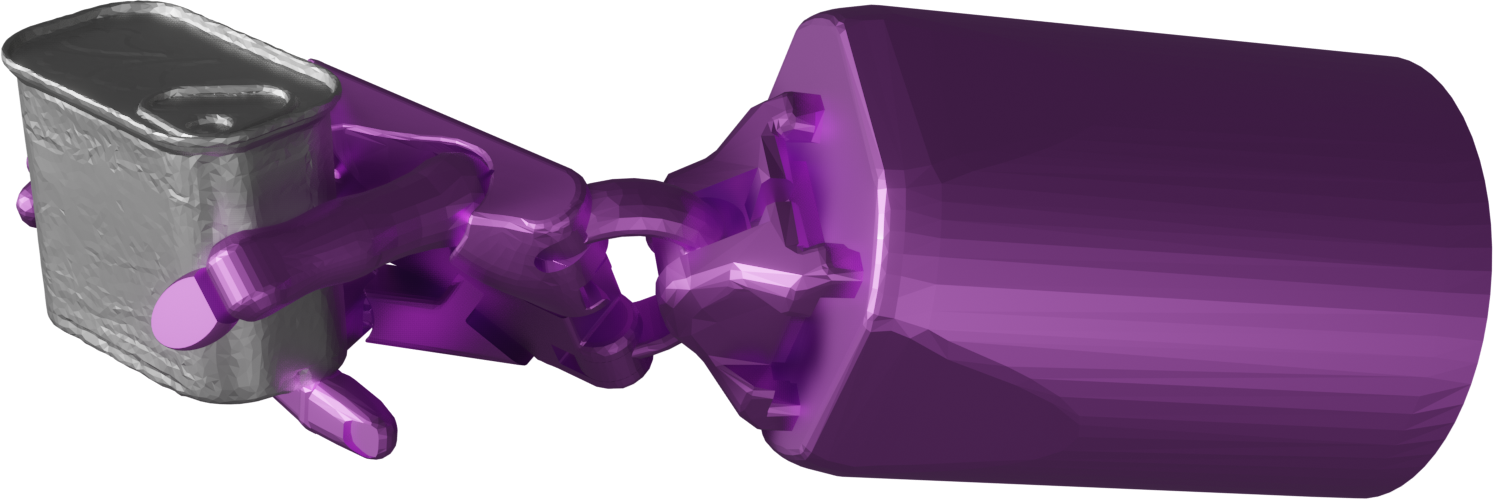}
    \hfill\end{subfigure} 
    \caption{\textbf{Exemplar grasps of different hands and objects from the proposed synthesized dataset.} From top to bottom: EZGripper, Barrett, Robotiq-3F, Allegro, and ShadowHand. From left to right: alarm clock, apple, binocular, and meat can.}
    \label{fig:dataset}
\end{figure}

\subsection{Grasp Datasets}

3D dexterous grasping poses are notoriously expensive to collect due to the complexity of hand structures. The industrial standard method of collecting a grasping pose is through kinesthetic demonstration~\cite{eiband2021identification}, wherein a human operator manually moves a physical robot towards a grasping pose. While researchers could collect high-quality demonstrations with kinesthetic demonstrations, it is considered too expensive for large-scale datasets. To tackle this challenge, researchers devised various low-cost data collection methods.

The straightforward idea is to replace kinesthetic demonstration with a motion capture system. Recent works have leveraged optical~\cite{taheri2020grab,taheri2022goal,fan2022articulated} and visual~\cite{hampali2020honnotate,hampali2022keypoint,chao2021dexycb,brahmbhatt2020contactpose} MoCap systems to collect human demonstrations. Another stream of work collects the contact map on objects by capturing the heat residual on the object surfaces after each human demonstration and using the contact map as a proxy for physical grasping hand pose~\cite{brahmbhatt2019contactdb,brahmbhatt2020contactpose}. Despite the differences in data collection pipelines, these prior arts collect human demonstrations within a limited setting, between pick-up and use. Such settings fail to cover the long-tail and complex nature of human grasping poses as depicted in the grasping taxonomy~\cite{feix2015grasp} and grasp landscape~\cite{liu2021synthesizing}. As a result, the collected grasping poses are similar to each other and can be represented by a few principal components~\cite{romero2017embodied,ciocarlie2007dimensionality}. We observe the same problem in programmatically generated datasets~\cite{goldfeder2009columbia,lundell2021multi,lundell2021ddgc,hasson2019learning,liu2020deep} using GraspIt!~\cite{miller2004graspit}.

\section{Dataset Collection}

To learn a versatile and hand-agnostic contact map generator, the grasp dataset ought to contain diverse grasping poses and corresponding contact maps for different objects and robotic hands with various morphologies. 

\subsection{Grasp Pose Synthesis}

Inspired by Liu \etal~\cite{liu2021synthesizing}, we synthesized a large-scale grasping dataset by minimizing a differentiable force closure estimator \dfc, a hand prior energy $E_\mathrm{n}$, and a penetration energy $E_\mathrm{p}$. We use the qpos $\qH$ to represent the kinematics pose of a robotic hand $H$, denoted as
\begin{equation}
    \qH = \{ q_\mathrm{global}\in\mathbb{R}^6, q_\mathrm{joint}\in\mathbb{R}^{N} \},
\end{equation}
where $q_\mathrm{global}$ is the rotation and translation of the root link, and $q_\mathrm{joint}$ describes the rotation angles of the revolute joints.

We selected \nobj daily objects from the YCB dataset~\cite{calli2017yale} and ContactDB~\cite{brahmbhatt2019contactdb}, together with 5 robotic hands (EZGripper, Barrett Hand, Robotiq-3F, Allegro, and Shadowhand) ranging from two to five fingers. We split our dataset into 48 training objects and 10 test objects. We show a random subset of the collected dataset in \cref{fig:dataset}.

\begin{figure}
    \centering
    \begin{minipage}{0.3\linewidth}
    \includegraphics[width=\linewidth]{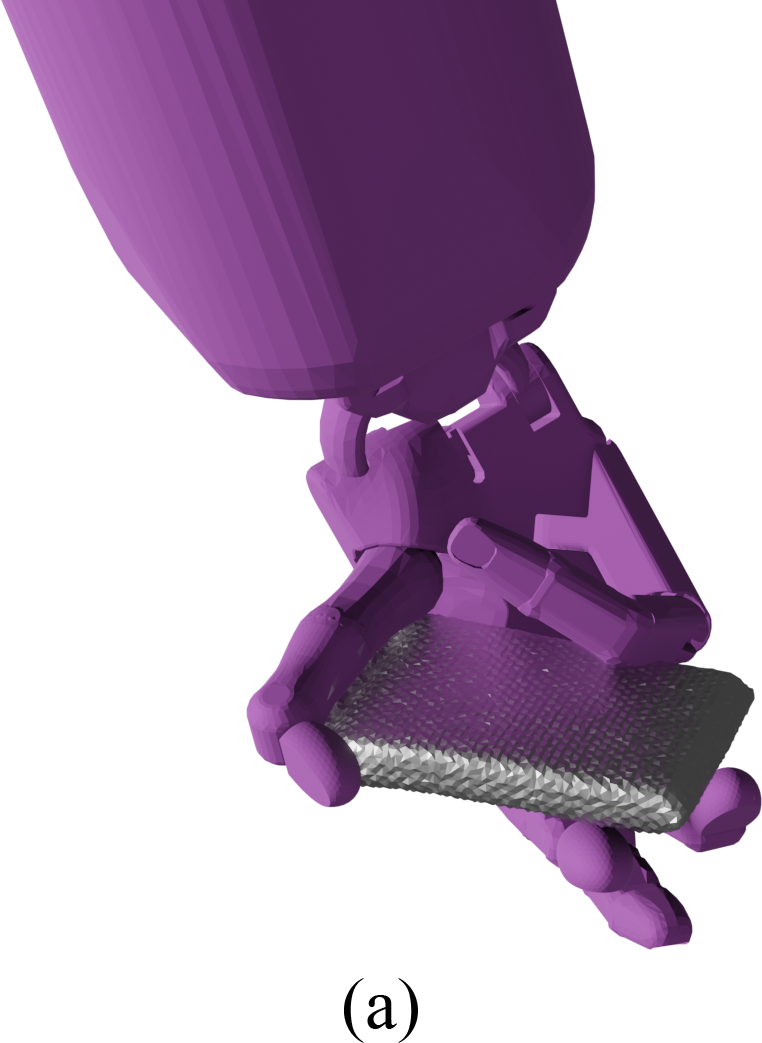}
    \end{minipage} \hfill
    \begin{minipage}{0.6\linewidth}
    \includegraphics[width=0.45\linewidth]{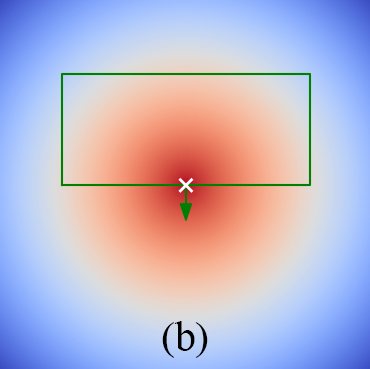}
    \includegraphics[width=0.45\linewidth]{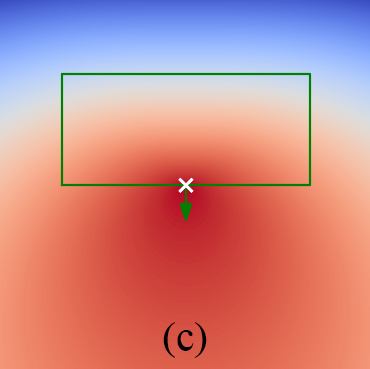}\vspace{0.3cm}\\
    \includegraphics[width=0.45\linewidth]{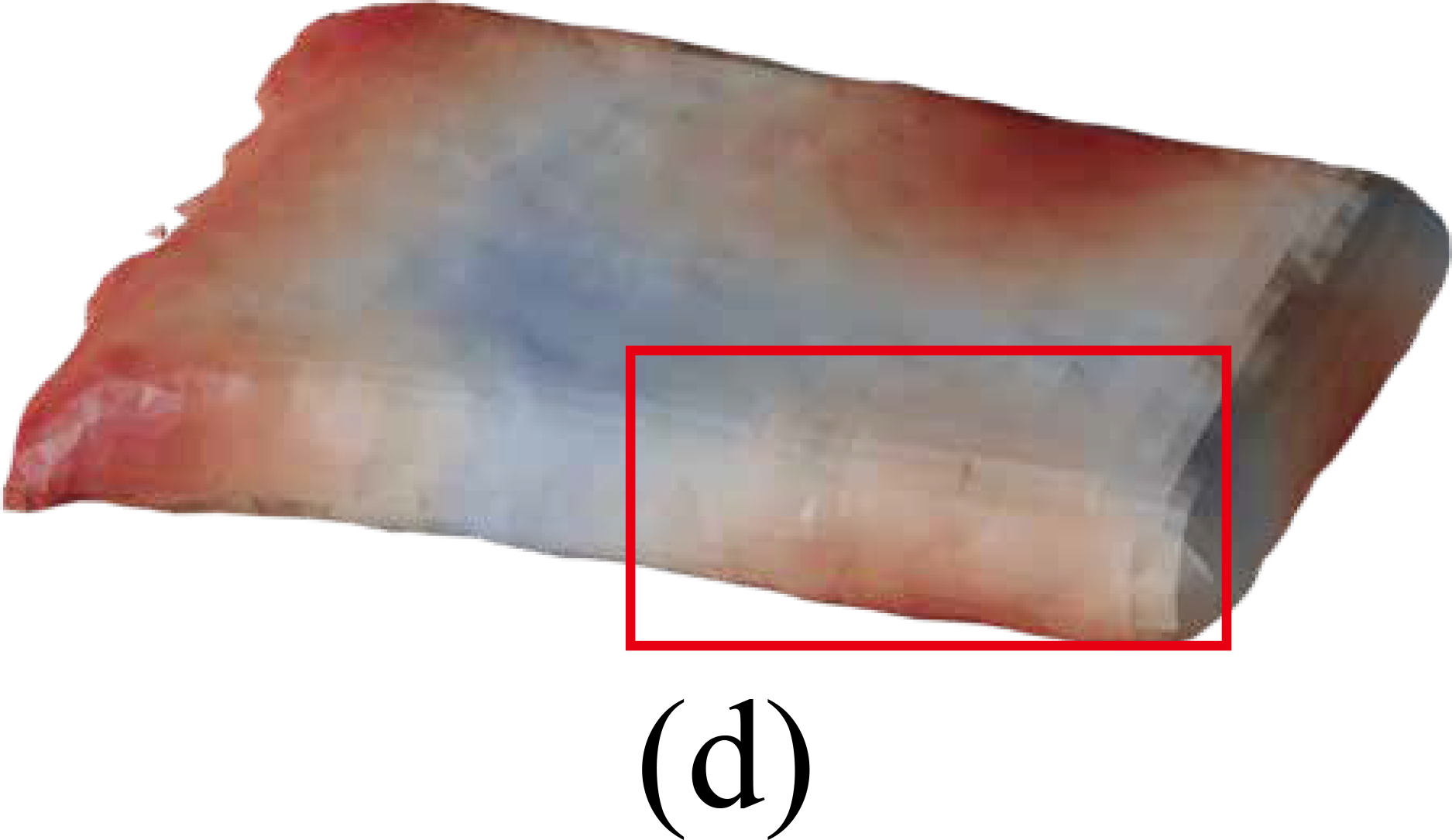}
    \includegraphics[width=0.45\linewidth]{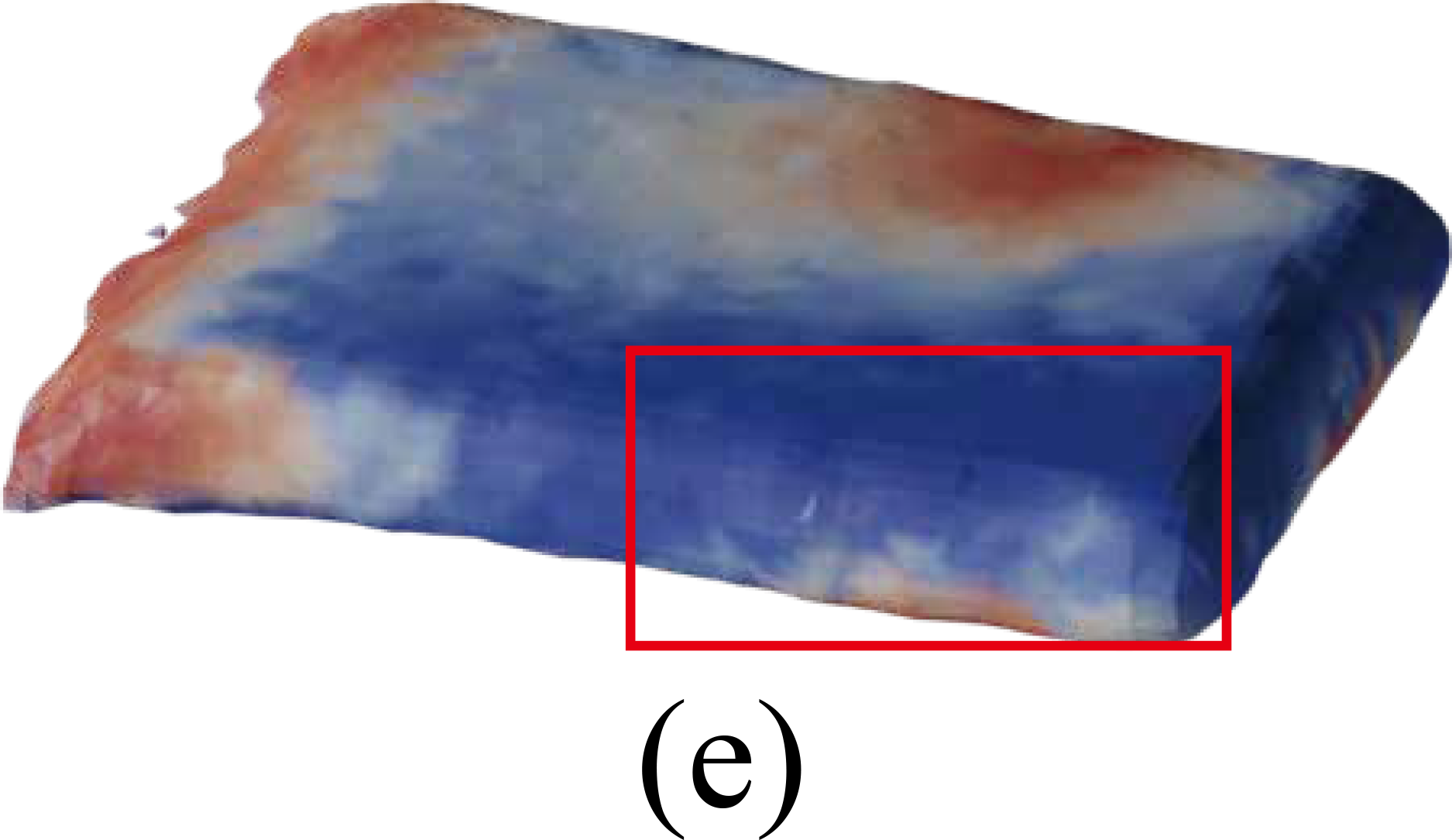}
    \end{minipage}
    \caption{\textbf{Comparison between aligned and euclidean distances on thin shell objects.} Given an exemplar grasp (a), we show both distances from the object to hand surfaces in 3D; red regions denote shorter distances and blue longer. An illustration of both distances is also shown in 2D (b,c); the green rectangle, white cross, and green arrow represent a rectangular object, the point of interest, and the surface normal $n_o$ at the point, respectively. The Euclidean distance (b) labels the upper edge of the object as close to the point of interest, whereas the aligned distance (c) is geometry-aware. The 3D aligned distances of the exemplar grasp (e) correctly reflect non-contact areas in the highlighted area, where the finger contacts the opposite side of the thin object. The Euclidean distances (d) fail to distinguish contacts on one side from contacts on the other side.}
    \label{fig:aligned_distance}
\end{figure}

Given an object $O$, a kinematics model of a robotic hand $H$ with pose $\qH$ and surface $\HH$, and a group of $n$ hand-centric contact points $X\subset\HH$, we define the differentiable force closure estimator \dfc as:
\begin{equation}
    \mathrm{\dfc} = Gc,
\end{equation}
where $c\in\mathbb{R}^{3n\times1}$ is the object surface normal on the contact points $X$, and
\begin{align}
    G &= \begin{bmatrix}
    I_{3\times 3} & I_{3\times 3} & ... & I_{3\times 3}\\
    \lfloor x_1 \rfloor_\times & \lfloor x_2 \rfloor_\times & ... & \lfloor x_n \rfloor_\times
    \end{bmatrix}, \label{eq:G}\\
    \lfloor x_i \rfloor_\times &= \begin{bmatrix}
    0 & -x_i^{(3)} & x_i^{(2)}\\
    x_i^{(3)} & 0 & -x_i^{(1)}\\
    -x_i^{(2)} & x_i^{(1)} & 0
    \end{bmatrix}.
\end{align}

\dfc describes the total wrench when each contact point applies equal forces, and friction forces are neglectable. As established in Liu \etal~\cite{liu2021synthesizing}, \dfc is a strong estimator of the classical force closure metric. 

Next, we define the prior and penetration energy as
\begin{align}
    E_\mathrm{p}(\qH, O) &= \sum_{x\in \HH} {\mathrm{R}(-\delta(x, O))} \label{eq:pen}\\
    E_\mathrm{n}(\qH) &= \lVert \mathrm{R}(\qH - {\qH}_\uparrow) + \mathrm{R}({\qH}_\downarrow - \qH)\rVert_2,
    \label{eq:norm}
\end{align}
where ${\qH}_\uparrow$ and ${\qH}_\downarrow$ are the upper and lower limits of the robotic hand parameters, respectively. $\delta(x, O)$ gives the signed distance from $x$ to $O$, where the distance is positive if $x$ is outside $O$ and is negative if inside.

Generating valid grasps requires finding the optimal set of contact points $X\subset\HH$ that minimize $E=\mathrm{\dfc}+E_\mathrm{n}+E_\mathrm{p}$. For computational efficiency, we sample $X\subset\HH$ from a set of rectangular contact regions predefined for each robotic hand. This strategy allows us to update the contact point positions via a gradient-based optimizer and improve sample efficiency. We use the DeepSDF~\cite{park2019deepsdf,davies2020overfit} to approximate the signed distance and surface normal of an object.

We use a Metropolis-adjusted Langevin algorithm (\mala)~\cite{liu2021synthesizing} to simultaneously sample the grasping poses and contact points. We run the \mala algorithm on an NVIDIA A100 80GB with a batch size of 1024 for each hand-object pair and obtain \ngrasp valid grasping poses. It takes about 1,400 GPU hours to synthesize the entire dataset.

\begin{figure*}[t!]
    \centering
    \includegraphics[width=\linewidth]{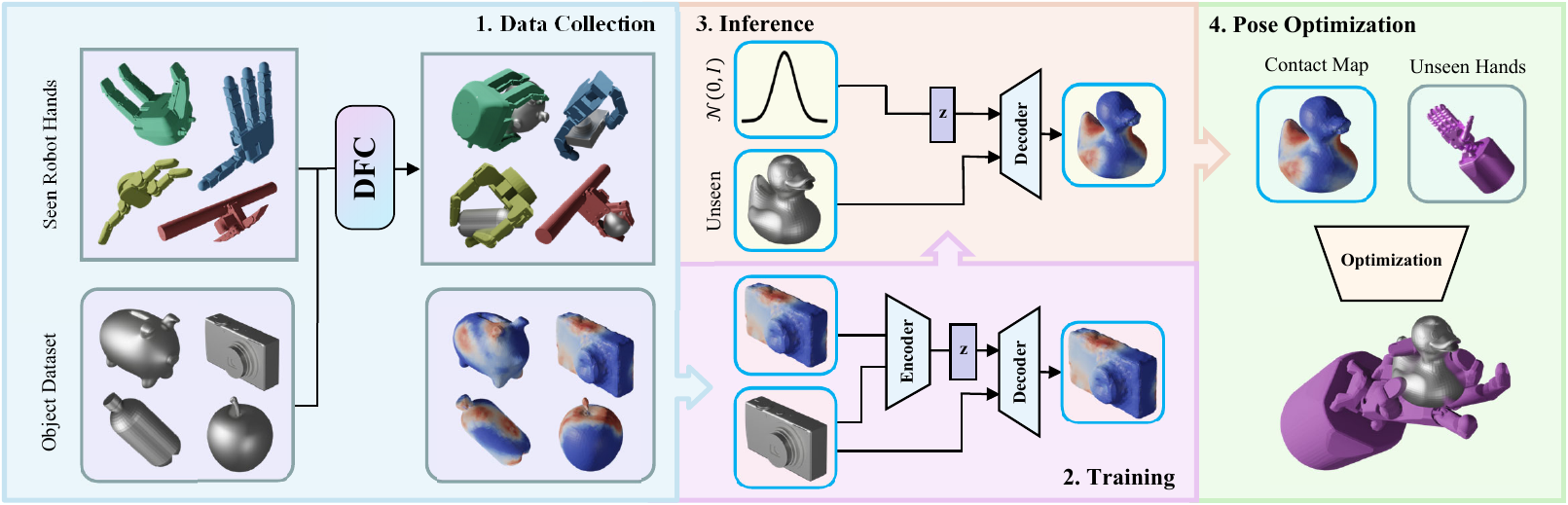}
    \caption{\textbf{An overview of the \method pipeline.} We first collect a large-scale synthetic dataset for multiple hands with \dfc. Then, we train a CVAE to generate hand-agnostic contact maps for unseen objects. We finally optimize grasping poses for unseen hands using the generated contact maps.}
    \label{fig:pipeline}
\end{figure*}

\subsection{Contact Map Synthesis}

Given the grasping poses, we first compute the object-centric contact map $\Omega$ as a set of normalized distances from each object surface point to the hand surface. Instead of using Euclidean distance, we propose an aligned distance to measure the distance between the object's surface point and the hand surface. Given the object $O$ and the hand $H$ with optimized grasp pose $\qH$, we define $\OO$ as the surface of $O$ and $\HH$ as the surface of $H$. The aligned distance $\DD$ between an object surface point $v_o\in\OO$ and $\HH$ is defined as:
\begin{equation}
    \DD(v_o, \HH) = \min_{v_h\in\HH} e^{\gamma(1-\langle v_o-v_h, n_o\rangle)} \sqrt{\Vert v_o - v_h \Vert_2},
    \label{eq:aligned_distance}
\end{equation}
where $\langle\cdot,\cdot\rangle$ denotes the inner product of two normalized vectors, and $n_o$ denotes the object surface normal at $v_o$. $\gamma$ is a scaling factor; we empirically set it to 1. The aligned distance considers directional alignment with the object's surface normal on the contact point and reduces contact ambiguities on thin-shell objects. \cref{fig:aligned_distance} shows that our aligned distance correctly distinguishes contacts from different sides of a thin shell, whereas the Euclidean distance mistakenly labels both sides as contact regions. 

Next, we compute the contact value $\CC(v_o,\HH)$ on each object surface point $v_o$ following Jiang \etal~\cite{jiang2021hand}:
\begin{equation} 
    \CC(v_o, \HH) = 1 - 2 \Big(\mathrm{Sigmoid}\big(\DD(v_o, \HH)\big) - 0.5\Big), 
    \label{eq:cv}
\end{equation}
where $\CC(v_o, \HH)\in(0,1]$ is $1$ if $v_o$ is in contact with $\HH$, and is $0$ if it is far away. $\CC \leq 1$ since $\DD$ is non-negative.

Finally, we define the contact map $\Omega(\OO,\HH)$ as
\begin{equation}
    \Omega(\OO, \HH) = \{ \CC(v_o, \HH) \}_{v_o \in \OO}.
    \label{eq:omega}
\end{equation}

\section{\method}

Given an object $O$ and the kinematics model of an arbitrary robotic hand $H$ with $N$ joints, we aim to generate a dexterous, diverse, and physically stable grasp pose $\qH$.

\subsection{Generate Hand-Agnostic Contact Maps}\label{sec:contact_map}

Generating $\qH$ directly for unseen $H$ is challenging due to the sparsity of the observed hands and the non-linearity between $\qH$ and hand geometry. Inspired by Brahmbhatt \etal~\cite{brahmbhatt2019contactgrasp}, we adopt the object-centric contact map as a hand-agnostic intermediate representation of a grasp. Instead of directly generating $\qH$, we first learn a generative model that generates a contact map over the object surface. We then fit the hand to the generated map.

Inspired by the successful applications of generative models in grasping~\cite{mousavian20196,taheri2020grab,jiang2021hand}, we adopt CVAE~\cite{sohn2015learning} to generate the hand-agnostic contact map. Given the point cloud of an input object and the corresponding pointwise contact values $\CC$, we use a PointNet~\cite{qi2017pointnet} encoder to extract the latent distribution $\mathcal{N}(\mu,\sigma)$ and sample the latent code $z\sim\mathcal{N}(\mu,\sigma)$. When decoding, we extract the object point features with another PointNet, concatenate $z$ to the per-point features, and use a shared-weight MLP to generate a contact value $\hat\CC(v_o)$ for each $v_o\in\OO$, which forms the predicted contact map $\hat\Omega(\OO)=\{\hat\CC(v_o)\}_{v_o\in\OO}$.

We learn the generative model by maximizing the log-likelihood of $p_{\theta, \varphi}(\Omega\mid O)$, where $\theta$ and $\phi$ are the learnable parameters of the encoder and decoder, respectively. According to Sohn \etal~\cite{sohn2015learning}, we equivalently maximize the ELBO:
\begin{equation}
    \begin{aligned}
        \log p_{\theta, \varphi}(\Omega\mid O) \geq & \mathbb{E}_{z\sim Z}[\log p_\varphi(\Omega\mid z, O)] \\
        & - D_{KL}(p_\theta(z\mid\Omega, O) \mid\mid p_Z(z)),
    \end{aligned}
\end{equation}
where $Z$ is the prior distribution of the latent space; we treat $Z$ as the standard normal distribution $\mathcal{N}(0, I)$. 

We leverage a reconstruction loss to approximate the expectation term of ELBO:
\begin{equation}
    \mathbb{E}_{z\sim Z}[\log p_\varphi(\Omega\mid z, O)] = \frac{1}{N_o}\sum_{i=0}^{N_o-1}{\| \hat{\Omega}^i - \Omega^i \|_2},
\end{equation}
where $N_o$ is the number of examples. $\Omega^i$ and $\hat\Omega^i$ denote the expected and generated contact map of the $i$-th example, respectively. 

Of note, since the generated contact map is empirically more ambiguous than the ground-truth contact map, we sharpen the generated contact map with
\begin{align}
    \hhO = 
    \begin{cases}
    \hat\Omega & \mathrm{if~} \hat\Omega < 0.5 \\
    1 & \mathrm{otherwise.}
    \end{cases}
\end{align}

\begin{figure*}
    \centering
    \begin{subfigure}{0.10\linewidth}\hfill
    \includegraphics[width=0.9\linewidth,valign=c]{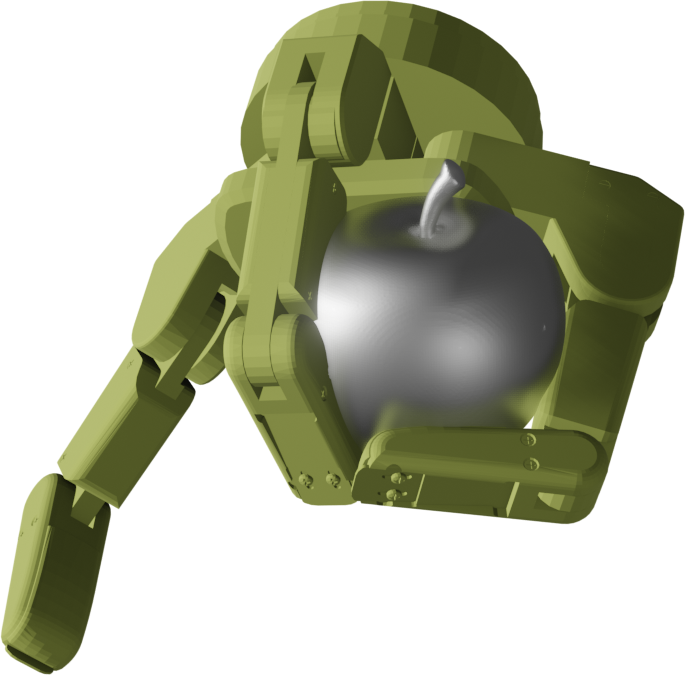} 
    \hfill\end{subfigure} \hfill
    \begin{subfigure}{0.10\linewidth}\hfill
    \includegraphics[width=0.9\linewidth,valign=c]{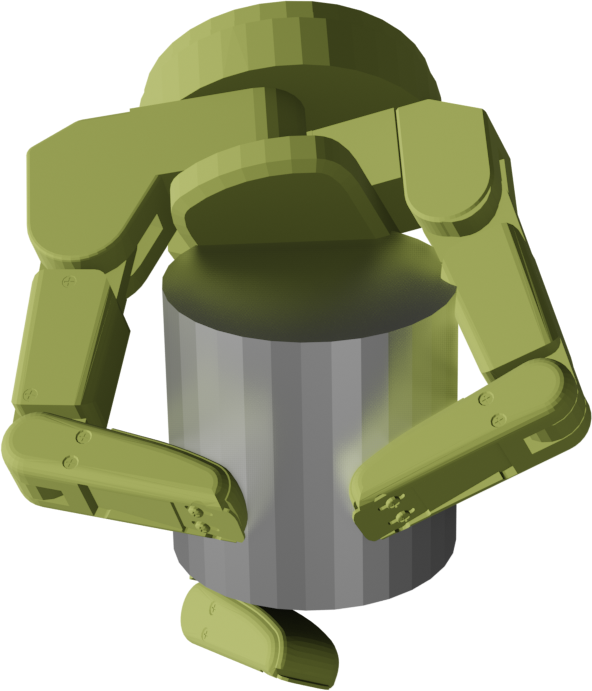} 
    \hfill\end{subfigure} \hfill
    \begin{subfigure}{0.10\linewidth}\hfill
    \includegraphics[width=0.7\linewidth,valign=c]{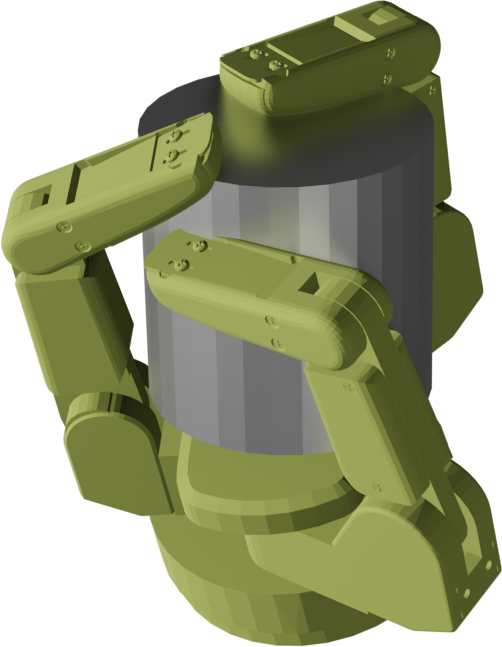} 
    \hfill\end{subfigure} \hfill
    \begin{subfigure}{0.10\linewidth}\hfill
    \includegraphics[width=0.9\linewidth,valign=c]{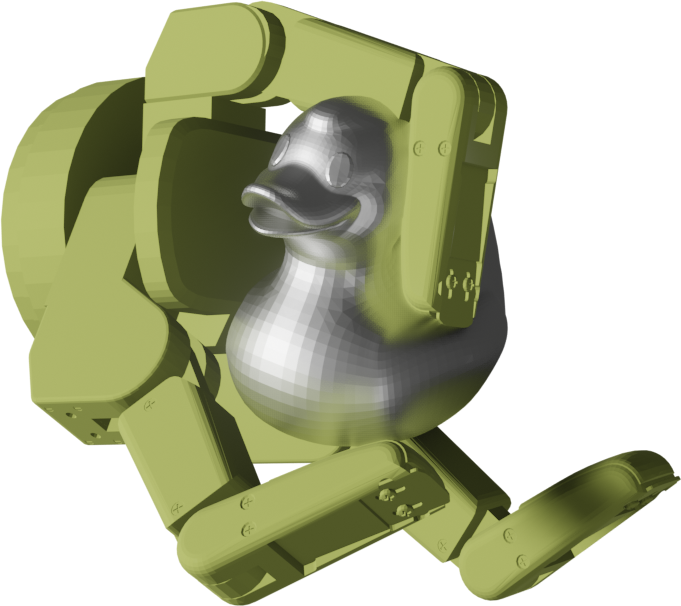} 
    \hfill\end{subfigure} \hfill
    \begin{subfigure}{0.10\linewidth}\hfill
    \includegraphics[width=0.9\linewidth,valign=c]{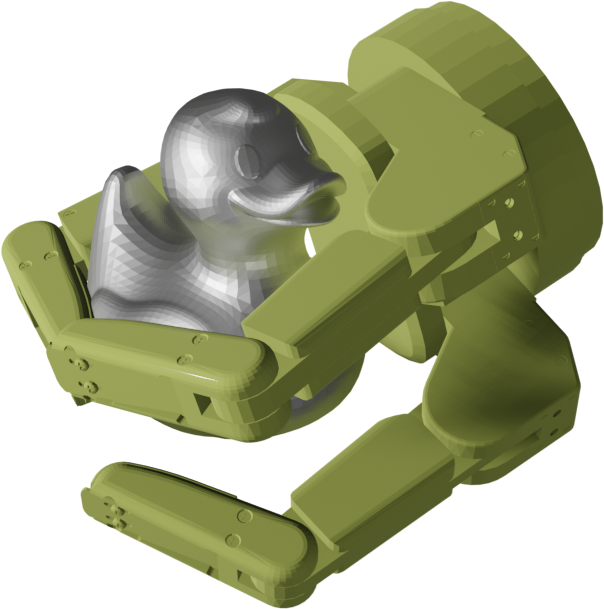} 
    \hfill\end{subfigure} \hfill
    \begin{subfigure}{0.10\linewidth}\hfill
    \includegraphics[width=0.9\linewidth,valign=c]{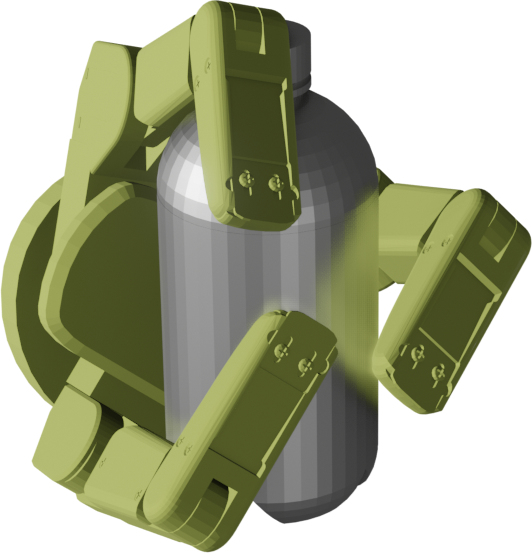} 
    \hfill\end{subfigure} \hfill
    \begin{subfigure}{0.10\linewidth}\hfill
    \includegraphics[width=0.9\linewidth,valign=c]{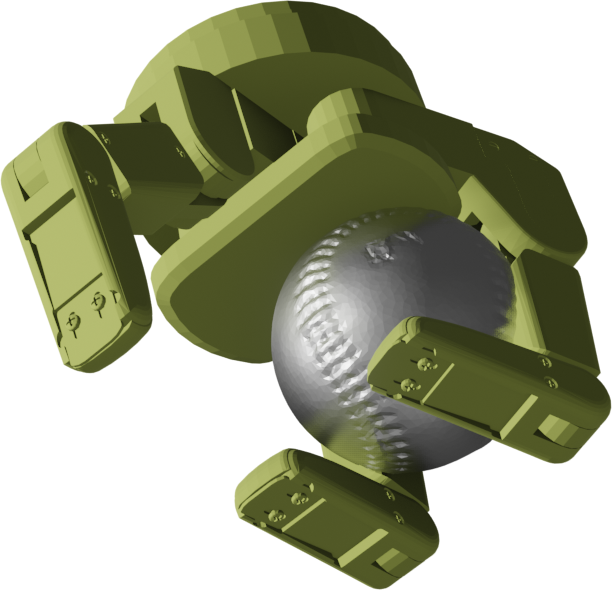} 
    \hfill\end{subfigure} \hfill
    \begin{subfigure}{0.10\linewidth}\hfill
    \includegraphics[width=0.9\linewidth,valign=c]{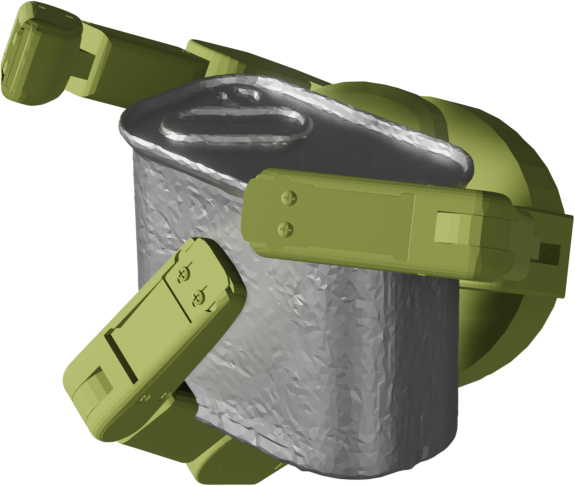} 
    \hfill\end{subfigure} 
    \\
    \begin{subfigure}{0.10\linewidth}\hfill
    \includegraphics[width=0.9\linewidth,valign=c]{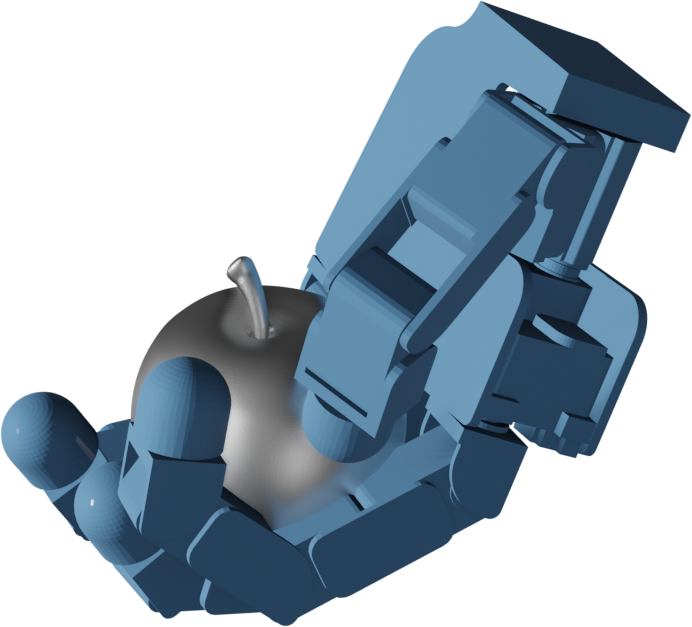} 
    \hfill\end{subfigure} \hfill
    \begin{subfigure}{0.10\linewidth}\hfill
    \includegraphics[width=0.9\linewidth,valign=c]{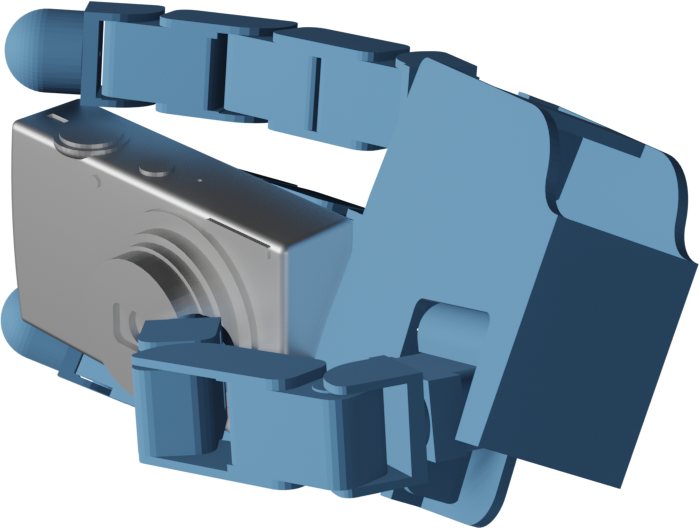} 
    \hfill\end{subfigure} \hfill
    \begin{subfigure}{0.10\linewidth}\hfill
    \includegraphics[width=0.7\linewidth,valign=c]{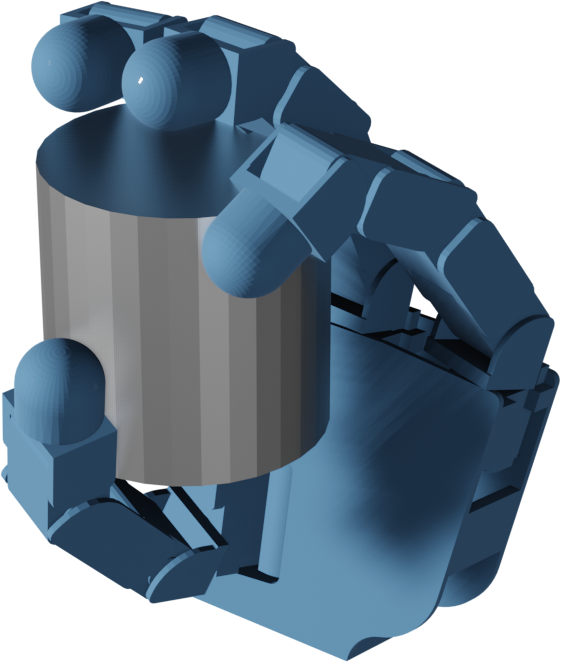} 
    \hfill\end{subfigure} \hfill
    \begin{subfigure}{0.10\linewidth}\hfill
    \includegraphics[width=0.9\linewidth,valign=c]{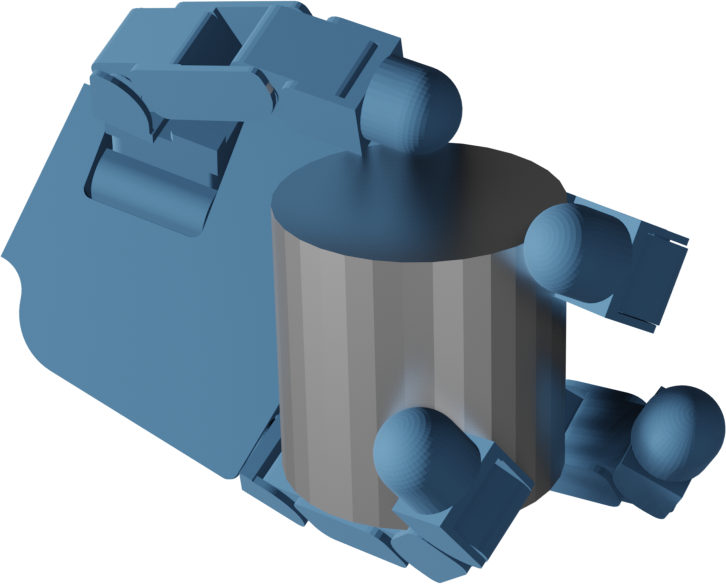} 
    \hfill\end{subfigure} \hfill
    \begin{subfigure}{0.10\linewidth}\hfill
    \includegraphics[width=0.9\linewidth,valign=c]{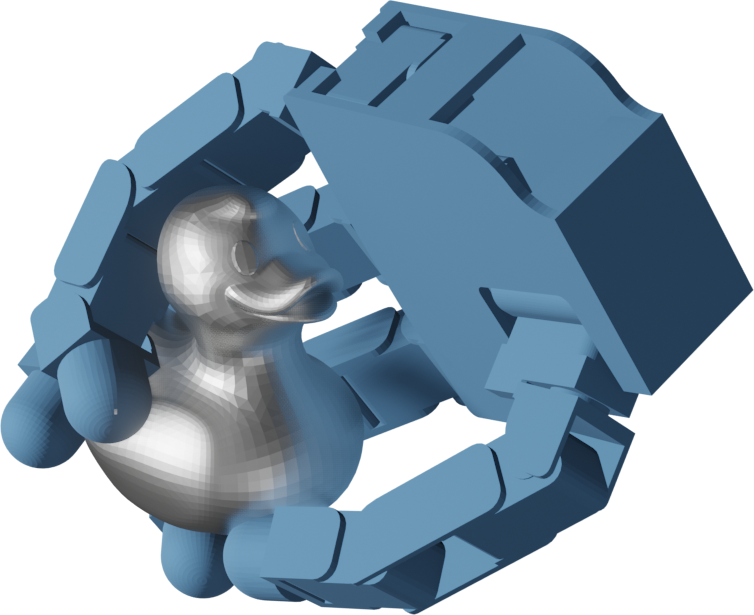} 
    \hfill\end{subfigure} \hfill
    \begin{subfigure}{0.10\linewidth}\hfill
    \includegraphics[width=0.9\linewidth,valign=c]{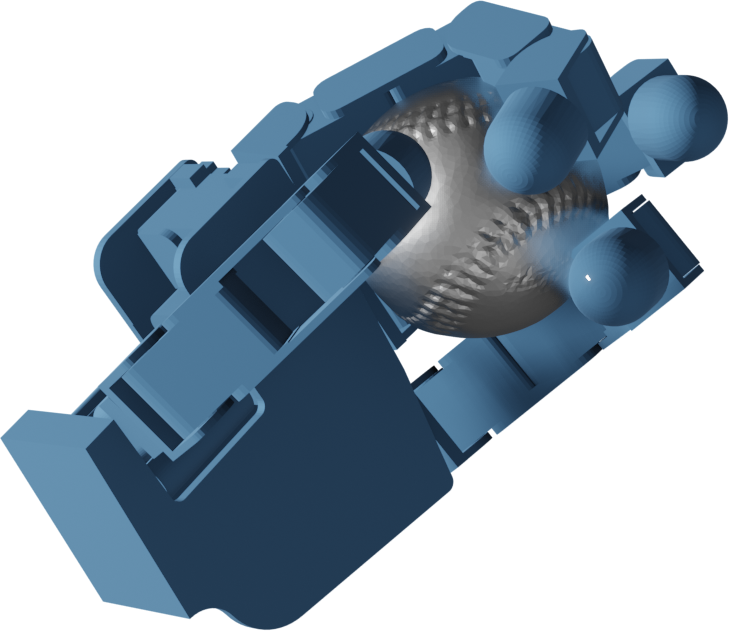} 
    \hfill\end{subfigure} \hfill
    \begin{subfigure}{0.10\linewidth}\hfill
    \includegraphics[width=0.9\linewidth,valign=c]{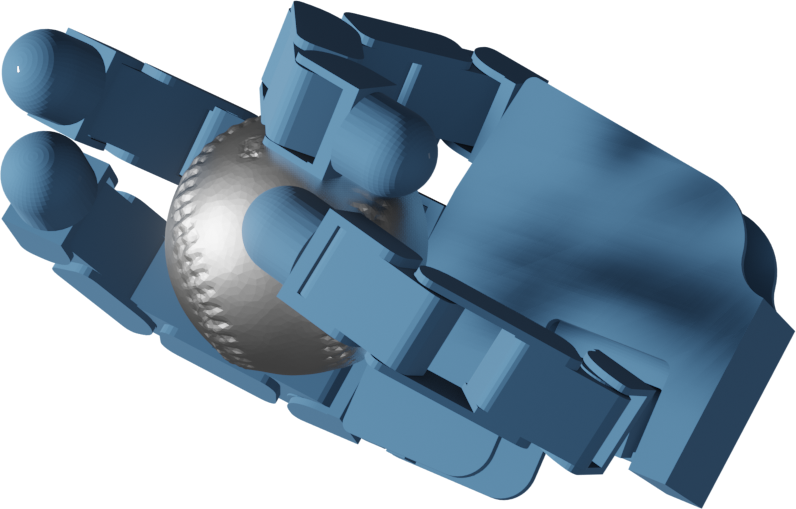} 
    \hfill\end{subfigure} \hfill
    \begin{subfigure}{0.10\linewidth}\centering
    \includegraphics[width=0.6\linewidth,valign=c]{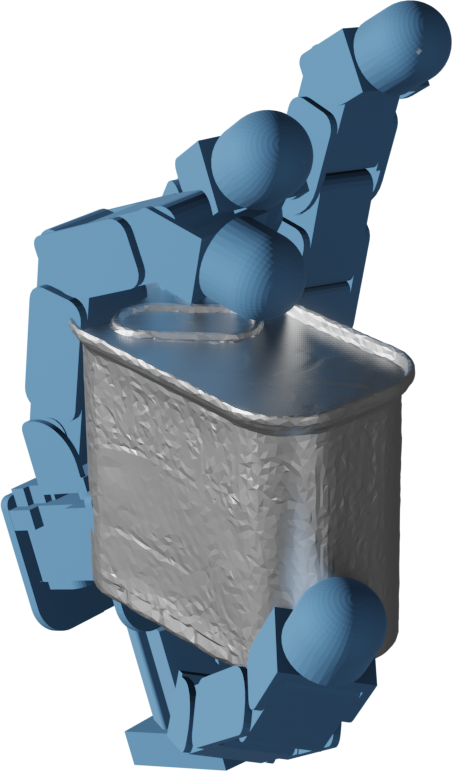} 
    \end{subfigure} 
    \\
    \begin{subfigure}{0.10\linewidth}\hfill
    \includegraphics[width=0.9\linewidth,valign=c]{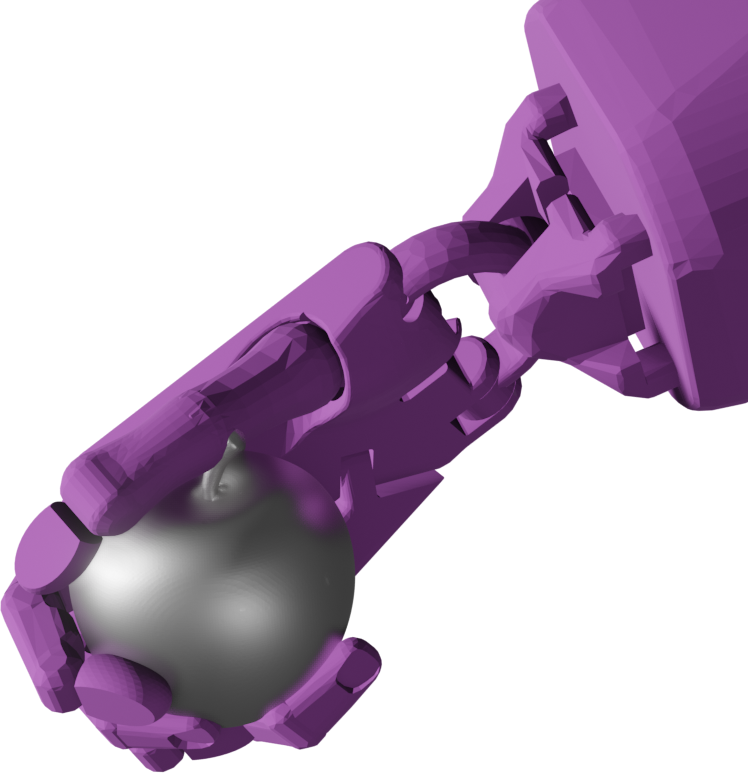} 
    \hfill\end{subfigure} \hfill
    \begin{subfigure}{0.10\linewidth}\hfill
    \includegraphics[width=0.9\linewidth,valign=c]{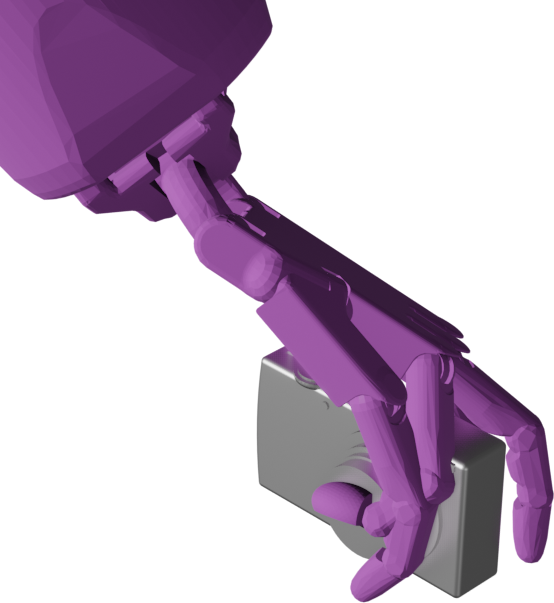} 
    \hfill\end{subfigure} \hfill
    \begin{subfigure}{0.10\linewidth}\hfill
    \includegraphics[width=0.9\linewidth,valign=c]{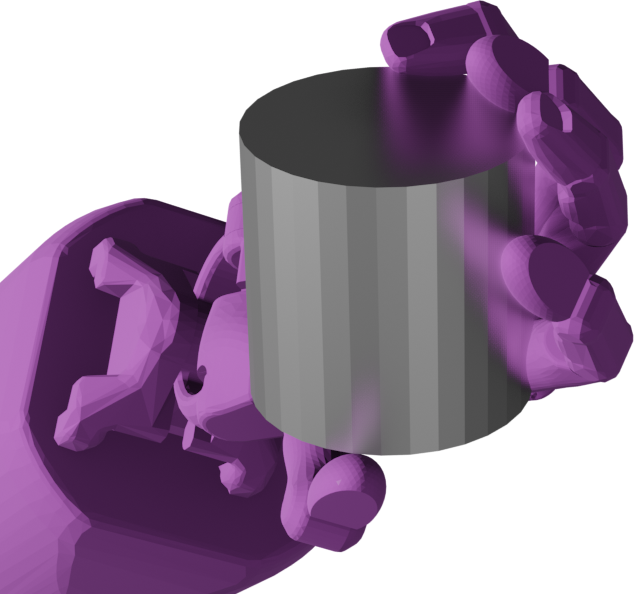} 
    \hfill\end{subfigure} \hfill
    \begin{subfigure}{0.10\linewidth}\hfill
    \includegraphics[width=0.9\linewidth,valign=c]{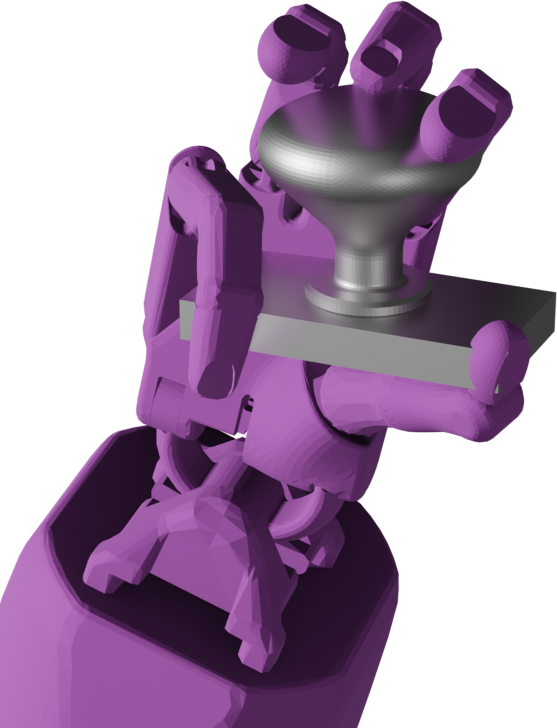} 
    \hfill\end{subfigure} \hfill
    \begin{subfigure}{0.10\linewidth}\hfill
    \includegraphics[width=0.9\linewidth,valign=c]{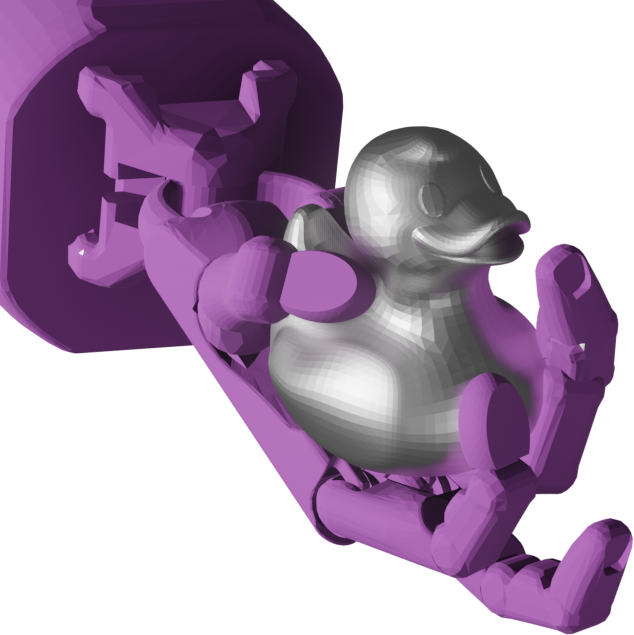} 
    \hfill\end{subfigure} \hfill
    \begin{subfigure}{0.10\linewidth}\hfill
    \includegraphics[width=0.6\linewidth,valign=c]{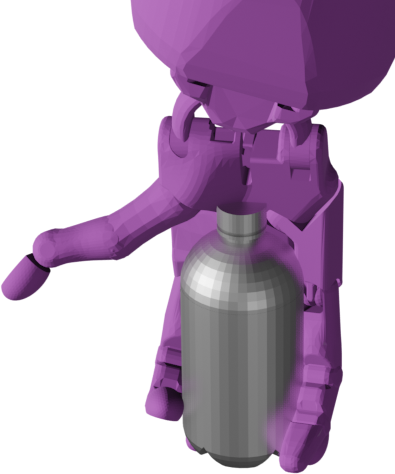} 
    \hfill\end{subfigure} \hfill
    \begin{subfigure}{0.10\linewidth}\hfill
    \includegraphics[width=0.9\linewidth,valign=c]{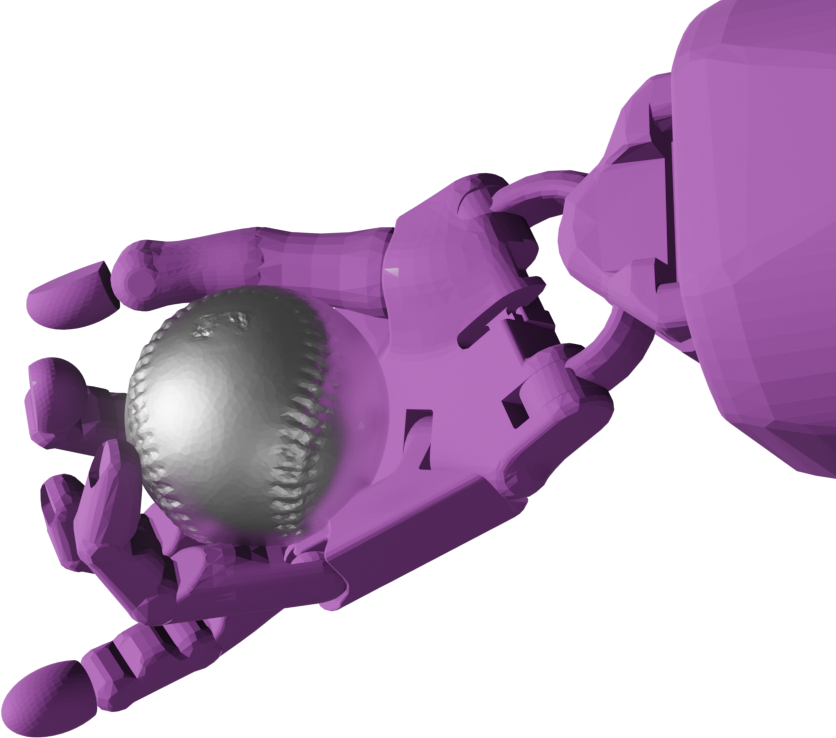}
    \hfill\end{subfigure} \hfill
    \begin{subfigure}{0.10\linewidth}\hfill
    \includegraphics[width=0.9\linewidth,valign=c]{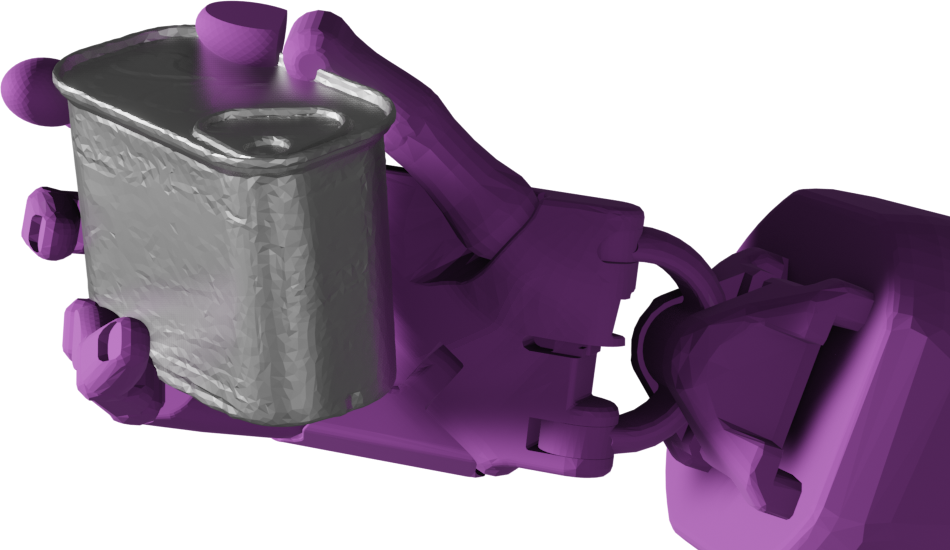} 
    \hfill\end{subfigure} 
    \\
    \caption{\textbf{Examples of the generated grasping poses for unseen hands and objects.} From top to bottom: Barrett, Allegro, and ShadowHand.}
    \label{fig:examples}
\end{figure*}

\subsection{Grasp Optimization}\label{sec:optim}

Given the generated contact map $\hhO$ on object $O$, we optimize the grasping pose $\qH$ for hand $H$. We initialize the optimization by randomly rotating the root link of the hand and translating the hand toward the back of its palm direction. We set the translation distance to the radius of the minimum enclosing sphere of the object. 

We compute $\HH$ by differentiable forward kinematics and obtain the current contact map $\dot\Omega$. We compute the optimization objective $E$ as
\begin{equation}
    E(\qH,\hhO,O)=E_\mathrm{c}(\qH,\hhO)+E_\mathrm{p}(\qH,O)+E_\mathrm{n}(\qH),
\end{equation}
where $E_\mathrm{c}$ is the MSE between the goal contact map $\hhO$ and the current contact map $\dot\Omega$. $E_\mathrm{p}$ and $E_\mathrm{n}$ describe the penetration between hand and object and if the hand pose is valid, respectively, described in \cref{eq:pen,eq:norm}.

Since the computation of the objective function is fully differentiable, we use the Adam optimizer to minimize $E$ by updating $\qH$. We run a batch of 32 parallel optimizations to keep the best result to avoid bad local minima.

\subsection{Implementation Details}

We optimize the CVAE for hand-agnostic contact maps using the Adam optimizer with a learning rate of $1\mathrm{e}{-4}$. Other Adam hyperparameters are left at default values. We train the CVAE for 36 epochs, which takes roughly 20 minutes on an NVIDIA 3090Ti GPU. The Adam optimizer for grasp uses a learning rate of $5\mathrm{e}{-3}$.

\section{Experiment}

\begin{figure}
    \centering
    \begin{subfigure}{0.22\linewidth}\hfill
    \includegraphics[width=0.8\linewidth,valign=c]{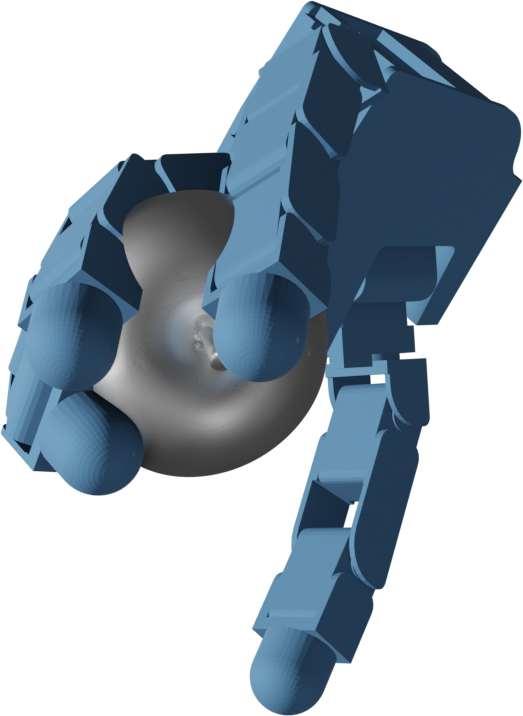} 
    \hfill\end{subfigure} \hfill
    \begin{subfigure}{0.22\linewidth}\hfill
    \includegraphics[width=0.9\linewidth,valign=c]{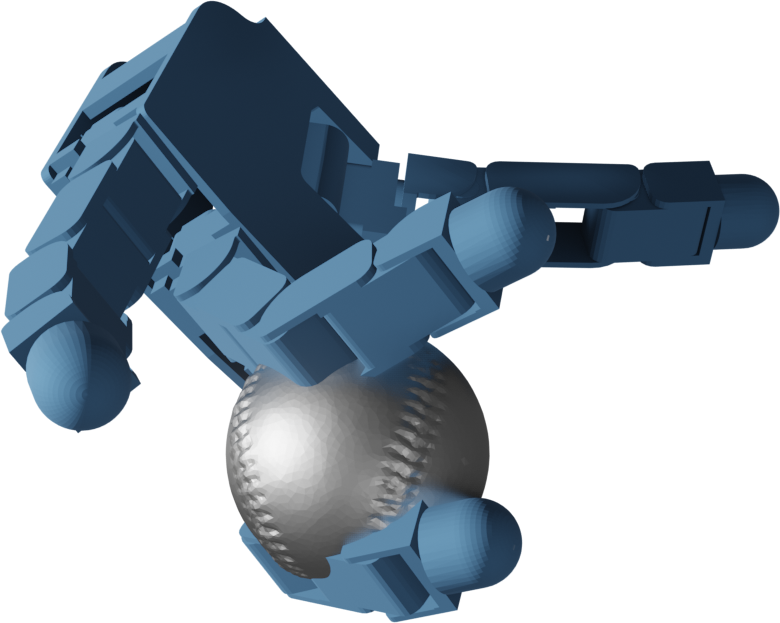} 
    \hfill\end{subfigure} \hfill
    \begin{subfigure}{0.22\linewidth}\hfill
    \includegraphics[width=0.8\linewidth,valign=c]{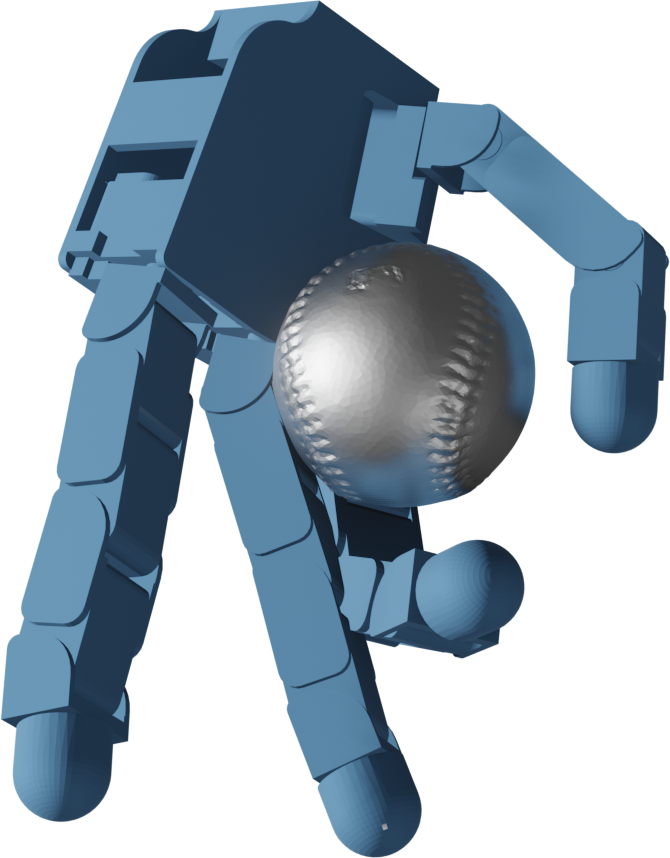} 
    \hfill\end{subfigure} \hfill
    \begin{subfigure}{0.22\linewidth}\centering
    \includegraphics[width=0.9\linewidth,valign=c]{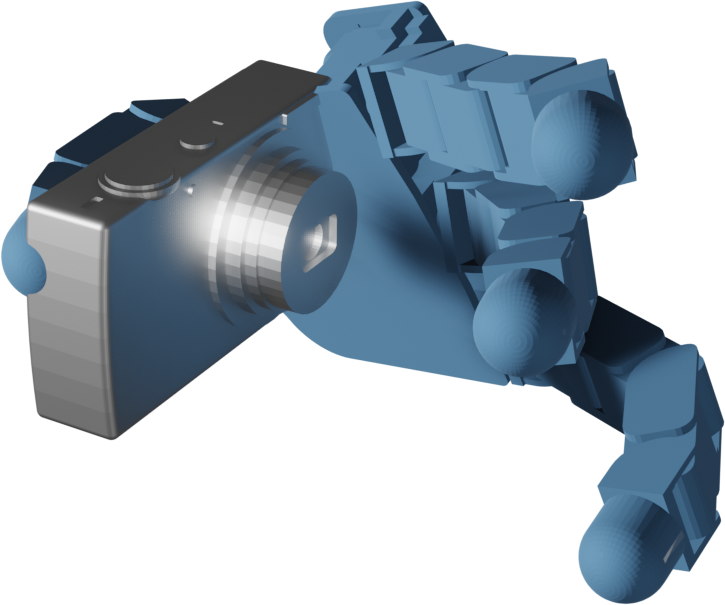} 
    \end{subfigure} 
    \\
    \begin{subfigure}{0.22\linewidth}\hfill
    \includegraphics[width=0.9\linewidth,valign=c]{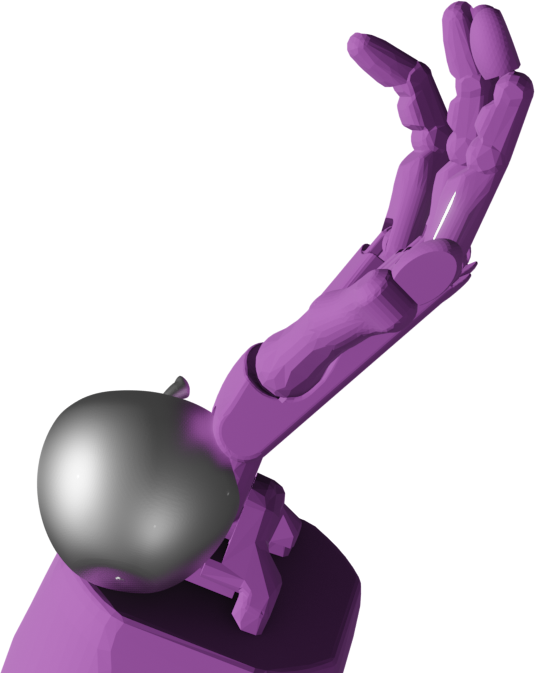} 
    \hfill\end{subfigure} \hfill
    \begin{subfigure}{0.22\linewidth}\hfill
    \includegraphics[width=0.9\linewidth,valign=c]{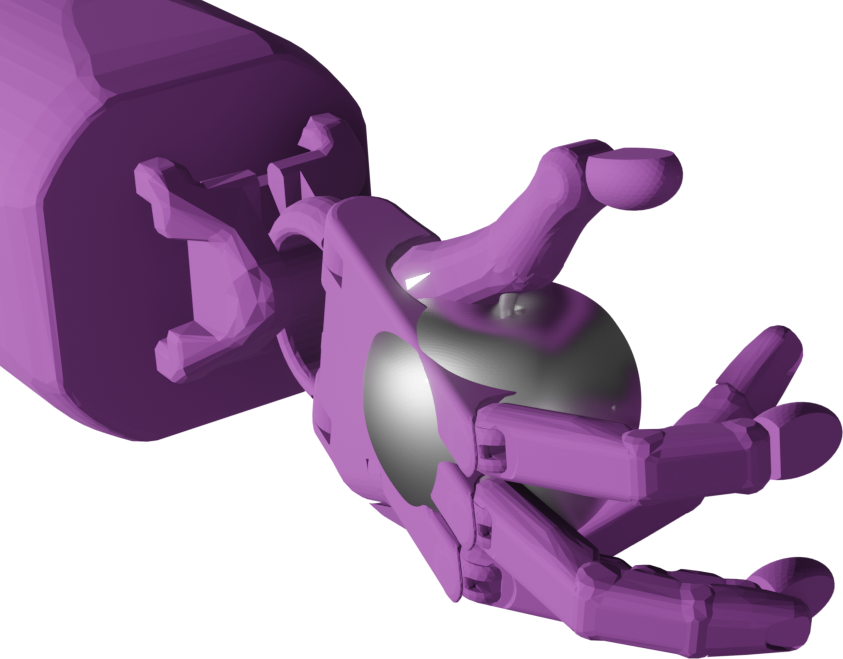} 
    \hfill\end{subfigure} \hfill
    \begin{subfigure}{0.22\linewidth}\hfill
    \includegraphics[width=0.7\linewidth,valign=c]{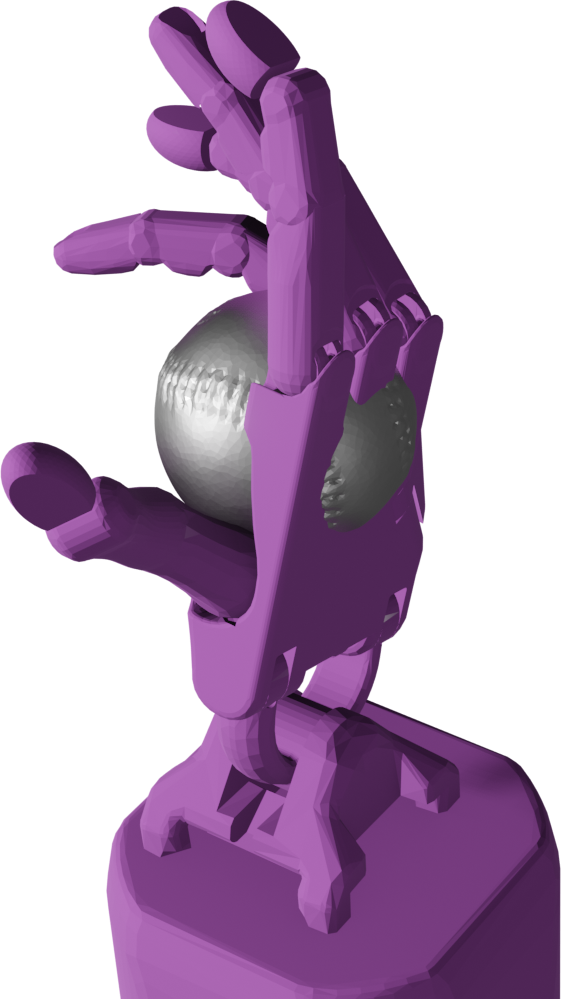} 
    \hfill\end{subfigure} \hfill
    \begin{subfigure}{0.22\linewidth}\hfill
    \includegraphics[width=0.9\linewidth,valign=c]{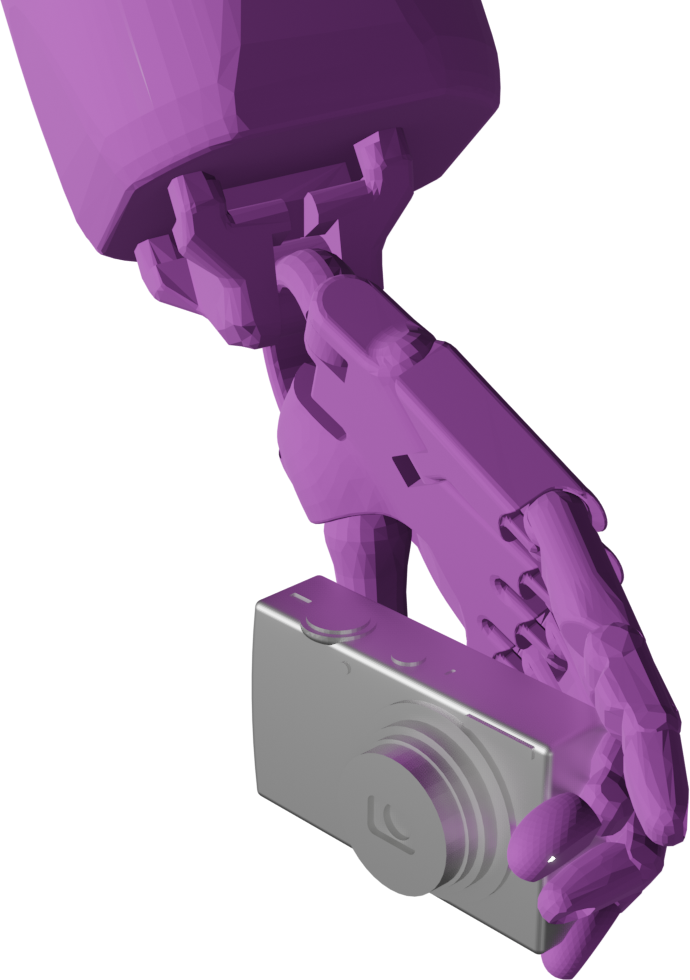} 
    \hfill\end{subfigure} 
    \\
    \caption{\textbf{Failure cases with Allegro (top) and ShadowHand (bottom).} The last column shows artifacts caused by contact ambiguities when using Euclidean distances instead of aligned distances.}
    \label{fig:failure}
\end{figure}

We quantitatively evaluate \method in terms of success rate, diversity, and inference speed. 

\paragraph*{Success Rate}

We test if a grasp is successful in the Isaac Gym environment~\cite{makoviychuk2021isaac} by applying an external acceleration to the object and measuring the movement of the object. We test each grasp by applying a consistent $0.5\mathrm{ms^{-2}}$ acceleration at the object for 1 second or 60 simulation steps and evaluate if the object moves more than 2cm after the simulation. We repeat this process for each grasp six times with acceleration along $\pm xyz$ directions. A grasp fails if it fails one of the six tests. Since generative methods usually exhibit minor errors that result in floatation and penetration near contact points, we apply a contact-aware refinement to the generated examples of all compared methods. Specifically, we first construct a target pose by moving the links close enough to the object (within 5mm) toward the object's direction. Next, we update $\qH$ with one step of gradient descent of step size 0.01 to minimize the difference between the current and the target pose. Finally, we track the updated pose with a positional controller provided by the Isaac Gym.

\paragraph*{Diversity}

We measure the diversity of the generated grasps as the standard deviation of the joint angles of the generated grasps that pass the simulation test. 

\paragraph*{Inference Speed}

We measure the time it takes for the entire inference pipeline to run.

\begin{table}[ht!]
    \centering
    \caption{Comparative Experiments}
    \label{tbl:comparison}
    \footnotesize
        \begin{tabular}{lcccc}
            \toprule
            \textbf{Methods} & \textbf{Gen.} & \textbf{Succ.$(\%)$} & \textbf{Div.$(\mathrm{rad.})$} & \textbf{Speed$(\mathrm{sec.})$} \\ \midrule
            \dfc \cite{liu2021synthesizing}  & \ding{51} & \textbf{79.53} & \textbf{0.344} & $>$1,800 \\ \midrule
            GC (w/o TTA)~\cite{jiang2021hand}   & \ding{55} & 19.38 & \textbf{0.340} & \textbf{0.012}  \\ 
            GC (w/ TTA)~\cite{jiang2021hand}    & \ding{55} & 22.03 & \textbf{0.355} & 43.233 \\ \midrule
            UniG.(top-1)~\cite{shao2020unigrasp} & \ding{51} & \textbf{80.00} & 0.000 & 9.331 \\ 
            UniG.(top-8)~\cite{shao2020unigrasp} & \ding{51} & 50.00 & 0.167 & 9.331 \\ 
            UniG.(top-32)~\cite{shao2020unigrasp} & \ding{51} & 48.44 & 0.202 & 9.331 \\ 
            \midrule
            Ours                            & \ding{51} & \textbf{77.19} & 0.207 & 16.415 \\ \bottomrule
        \end{tabular}%
\end{table}

We compare \method with \dfc~\cite{liu2021synthesizing}, GraspCVAE~\cite{jiang2021hand} (GC), and UniGrasp~\cite{shao2020unigrasp} (UniG.) in~\cref{tbl:comparison}. The columns represent method names, whether the method is generalizable, success rate, diversity, and inference speed. We evaluate all methods with the test split of the ShadowHand data in \dataset. We trained our method with the training split of EZGripper, Robotiq-3F, Barrett, and Allegro. Since GraspCVAE is designed for one specific hand structure, we train GraspCVAE on the training split of the ShadowHand data and keep the result before and after test-time adaptation (TTA). We evaluate UniGrasp with its pre-trained weights.

Of note, since the UniGrasp model only produces three contact points, we align them to the thumb, index, and middle finger of the ShadowHand for inverse kinematics. In addition, UniGrasp yields zero diversity since it produces the top-1 contact point selection for each object. To evaluate its diversity, we include top-8, top-32, and top-64 contact point selections. 
We observe that \dfc achieves the best success rate and diversity but is overwhelmingly slow. GraspCVAE can generate diverse grasping poses but suffers a low success rate and cannot generalize to unseen hands. We attribute the low success rate to our dataset's large diversity of grasping poses. The original GraspCVAE was trained on HO3D~\cite{hampali2020honnotate}, where grasp poses are similar since six principal components can summarize most grasping poses. 
UniGrasp can generalize to unseen hands and achieve a high success rate. However, it fails to balance success rate and diversity.

Our method achieves a slightly lower success rate than \dfc and UniGrasp top-1 but can generate diverse grasping poses in a short period of time, achieving an excellent three-way trade-off among quality, diversity, and speed.

We examine the efficacy of the proposed aligned distance in \cref{tbl:ablation-contact}. Specifically, we evaluate the success rate and diversity of the full model (full) and the full model with Euclidean distance contact maps (-align). The experiment is repeated on EZGripper, Barrett, and ShadowHand to show efficacy across hands. In all three cases, we observe that using the Euclidean distance lowers the success rate significantly while improving the diversity slightly. Such differences meet our expectations, as contact maps based on Euclidean distances are more ambiguous than those based on aligned distances. During the evaluation, such ambiguities bring more uncertainties, which are treated as diversities using our current metrics. We also observe that the model performs worse on the EZGripper due to the ambiguities in aligning two-finger grippers to multi-finger contact maps.

\begin{table}[ht!]
    \centering
    \caption{Ablation Study - Contact}
    \label{tbl:ablation-contact}
    \small
        \begin{tabular}{lcc}
            \toprule
            \textbf{Methods} & \textbf{Succ. Rate$(\%)$} & \textbf{Diversity$(\mathrm{rad.})$} \\ \midrule
            Full (EZGripper)        & \textbf{38.59} & 0.248 \\  
            -align (EZGripper)      & 29.53 & \textbf{0.312} \\ \midrule
            Full (Barrett)          & \textbf{70.31} & 0.267 \\ 
            -align (Barrett)        & 52.19 & \textbf{0.349} \\ \midrule
            Full (ShadowHand)       & \textbf{77.19} & 0.207 \\ 
            -align (ShadowHand)     & 58.91 & \textbf{0.237} \\ \bottomrule
        \end{tabular}%
\end{table}

\begin{figure}
    \centering
    \includegraphics[width=\linewidth]{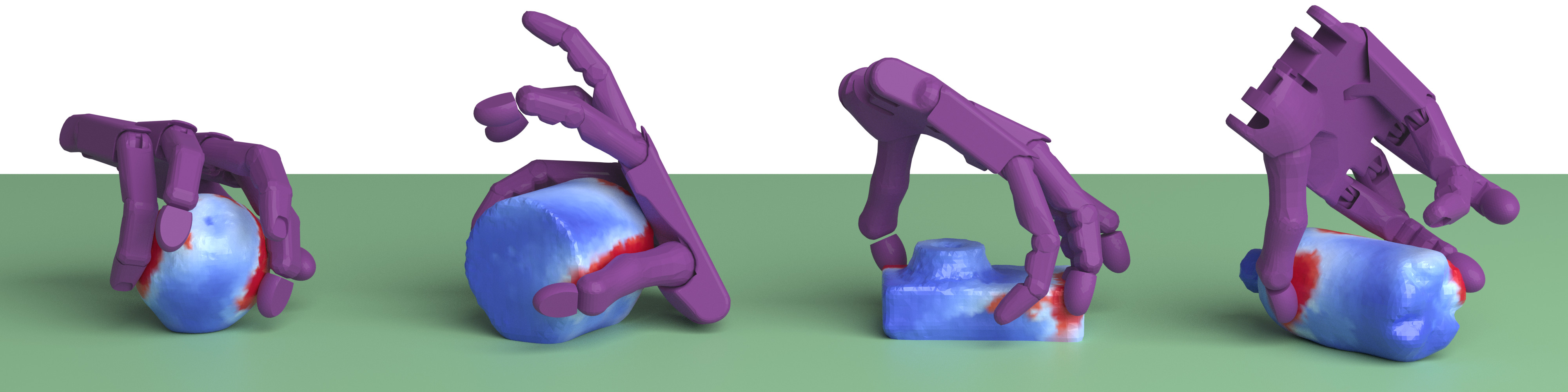}%
    \caption{\textbf{\method in tabletop scenarios.}}
    \label{fig:real}
\end{figure}

We further compare the performances of \method on seen and unseen hands in \cref{tbl:ablation-domain}. We train two versions of \method for each hand. The in-domain version is trained on all five hands and evaluated on the selected hand. The out-of-domain version is trained on all four hands except the selected hand and evaluated on the selected hand. Our result shows that our method is robust for various hand structures in out-of-domain scenarios. 

\begin{table}[ht!]
    \centering
    \caption{Ablation Study - Generalization}
    \label{tbl:ablation-domain}
    \small
        \begin{tabular}{lccc}
            \toprule
            \textbf{Robots} & \textbf{Domain} & \textbf{Succ. Rate$(\%)$} & \textbf{Diversity$(\mathrm{rad.})$} \\ \midrule
            Ezgripper    & in  & \textbf{43.44} & 0.238 \\  
            Ezgripper    & out & 38.59 & \textbf{0.248} \\ \midrule
            Barrett      & in  & \textbf{71.72} & \textbf{0.281} \\  
            Barrett      & out & 70.31 & 0.267 \\ \midrule
            Shadowhand   & in  & 77.03 & \textbf{0.211} \\  
            Shadowhand   & out & \textbf{77.19} & 0.207 \\ \bottomrule
        \end{tabular}%
\end{table}

The qualitative results in \cref{fig:examples} show the diversity and quality of grasps generated by \method. The generated grasps cover diverse grasping types, including wraps, pinches, tripods, quadpods, hooks, \etc. We also show failure cases in \cref{fig:failure}, where the first three columns show failures from our full model, and the last column shows failures specific to the \textit{-align} ablation version. The most common failure types are penetrations and floatations caused by imperfect optimization. We observe an interesting failure case in the first example in the bottom row, where the algorithm tries to grasp the apple by squeezing it between the palm and the base. While the example fails to pass the simulation test, it shows the level of diversity that our method provides. 

Finally, we demonstrate that our approach can be applied to tabletop objects after proper training; see \cref{fig:real}.

\section{Conclusion}

This paper introduces \method, a versatile dexterous grasping method that can generalize to unseen hands. By leveraging the contact map representation as the intermediate representation, a novel aligned distance for measuring hand-to-point distance, and a novel grasping algorithm, \method can generate diverse and high-quality grasping poses in reasonable inference time. The quantitative experiment suggests that our method is the first generalizable grasping algorithm to properly balance among quality, diversity, and speed. In addition, we contribute \dataset, a large-scale synthetic dexterous grasping dataset. \dataset features diverse grasping poses, a wide range of household objects, and five robotic hands with diverse kinematic structures.

\textbf{Acknowledgement:}
This work is supported in part by the National Key R\&D Program of China (2021ZD0150200), the Beijing Municipal Science \& Technology Commission (Z221100003422004), and the Beijing Nova Program.

\bibliographystyle{IEEEtran}
\balance
\bibliography{reference}
\end{document}